\documentclass[accepted]{article}

\usepackage{microtype}
\usepackage{graphicx}
\usepackage{subcaption}
\usepackage{booktabs} 
\usepackage{tabularx}
\usepackage{array}
\newcolumntype{L}[1]{>{\raggedright\arraybackslash}p{#1}}
\newcolumntype{C}[1]{>{\centering\arraybackslash}p{#1}}
\usepackage{tabularray}
\usepackage{multirow}
\usepackage{makecell}
\usepackage{arydshln}
\usepackage{amsmath}
\usepackage{graphicx}
\usepackage{import}
\usepackage{subcaption}
\usepackage[font=small,labelfont=bf]{caption}
\usepackage[export]{adjustbox}
\usepackage{pythonhighlight}

\usepackage[utf8]{inputenc} 
\usepackage[T1]{fontenc}    
\usepackage{url}            
\usepackage{booktabs}       
\usepackage{amsfonts}       
\usepackage{nicefrac}       
\usepackage{microtype}      
\usepackage{xcolor}         
\definecolor{lightgray}{gray}{0.95} 

\usepackage{hyperref}

\usepackage[accepted]{icml2026}

\usepackage{amsmath}
\usepackage{amssymb}
\usepackage{mathtools}
\usepackage{amsthm}

\usepackage[capitalize,noabbrev]{cleveref}

\theoremstyle{plain}

\theoremstyle{definition}

\theoremstyle{remark}

\newcommand{\R}{\mathbb{R}}

\usepackage[textsize=tiny]{todonotes}

\icmltitlerunning{PLAID: A Unified Data Model for Machine Learning on Heterogeneous Physics Simulations}

\begin{document}

\twocolumn[
  \icmltitle{PLAID: A Unified Data Model for Machine Learning on Heterogeneous Physics Simulations}



  \icmlsetsymbol{equal}{*}

  \begin{icmlauthorlist}
    \icmlauthor{Fabien Casenave}{safran}
    \icmlauthor{Xavier Roynard}{safran}
    \icmlauthor{Brian Staber}{safran}
    \icmlauthor{Alexandre Devaux-Rivi\`ere}{safran,ens}
    \icmlauthor{William Piat}{safran}
    \icmlauthor{Michele Alessandro Bucci}{safran}
    \icmlauthor{Nissrine Akkari}{safran}
    \icmlauthor{Abbas Kabalan}{safran,ponts}
    \icmlauthor{Xuan Minh Vuong Nguyen}{safran,mines}
    \icmlauthor{Luca Saverio}{safran,polytechnique,onera}
    \icmlauthor{Rapha\"{e}l Carpintero Perez}{safran,polytechnique}
    \icmlauthor{Anthony Kalaydjian}{safran,epfl}
    \icmlauthor{Samy Fouch\'{e}}{safran,ens}
    \icmlauthor{Thierry Gonon}{safran}
    \icmlauthor{Ghassan Najjar}{safran}
    \icmlauthor{Thomas Daniel}{safran}
    \icmlauthor{Emmanuel Menier}{augur}
    \icmlauthor{Matthieu Nastorg}{augur,inria}
    \icmlauthor{Giovanni Catalani}{airbus,supaero}
    \icmlauthor{Christian Rey}{safran}
  \end{icmlauthorlist}

  \icmlaffiliation{safran}{SafranTech, Paris, France}
  \icmlaffiliation{ponts}{Ecole des Ponts ParisTech (CERMICS), Paris, France}
  \icmlaffiliation{mines}{Mines Paris - PSL (CEMEF), Paris, France}
  \icmlaffiliation{polytechnique}{Ecole Polytechnique (CMAP), Paris, France}
  \icmlaffiliation{onera}{ONERA (DAAA), Paris, France}
  \icmlaffiliation{epfl}{EPFL, Lausanne, Switzerland}
  \icmlaffiliation{ens}{ENS Paris-Saclay, Paris, France}
  \icmlaffiliation{augur}{Augur, Paris, France}
  \icmlaffiliation{inria}{Inria, Paris, France}
  \icmlaffiliation{airbus}{Airbus, Toulouse, France}
  \icmlaffiliation{supaero}{ISAE-SUPAERO, Toulouse, France}

  \icmlcorrespondingauthor{Fabien Casenave}{fabien.casenave@safrangroup.com}

  \icmlkeywords{Physics Machine Learning, Benchmarking, Numerical Simulation, Data Model}

  \vskip 0.3in
]



\printAffiliationsAndNotice{}  

\begin{abstract}
Machine learning-based surrogate models have emerged as a powerful tool to accelerate simulation-driven scientific workflows, but their adoption is limited by the lack of large-scale, diverse, and standardized datasets for physics-based simulations. Existing benchmarks often focus on narrow domains or rely on simplified data models, and fail to capture the heterogeneity arising from variable geometries, meshes, and topologies, which is critical for assessing generalization in realistic settings.
We introduce PLAID (Physics-Learning AI Data model), a unified and extensible data layer for heterogeneous physics simulations. It preserves the full complexity of simulation data while enabling efficient and scalable machine learning workflows, together with a library for dataset construction and manipulation~(\href{https://github.com/PLAID-lib/plaid}{github.com/PLAID-lib/plaid}).
We release six datasets covering structural mechanics and computational fluid dynamics, designed to reflect realistic industrial scenarios and provide standardized benchmarks.
The framework includes reproducible evaluation protocols and is integrated with Hugging Face to enable open, community-driven benchmarking with active user participation (\href{https://huggingface.co/PLAIDcompetitions}{huggingface.co/PLAIDcompetitions}).
\end{abstract}

\section{Introduction}


Numerical simulation is central to scientific and engineering research, providing insights into complex physical phenomena across many domains~\cite{zienkiewicz1971finite, zienkiewicz2005finite, press2007numerical, samanidou2007agent, viceconti2016virtual, brotzge2023challenges}. These simulations typically rely on solving partial differential equations using large-scale numerical solvers, and are often computationally expensive, with single high-fidelity runs requiring hours or days. In many-query settings such as design exploration, optimization, or uncertainty quantification, this cost becomes prohibitive. Surrogate modeling techniques have therefore been developed to approximate simulation outputs at a fraction of the cost.

Classical surrogate models perform non-linear regression over parametric spaces using statistical learning techniques, such as polynomial regression, nearest neighbors, support vector machines, random forests~\cite{breiman2001random}, and Gaussian processes~\cite{williams2006gaussian}. These models are widely supported by software libraries such as UQLab~\cite{uqlab}, OpenTURNS~\cite{openturns}, Dakota~\cite{dakota2024} and Lagun~\cite{lagun}. However, they are typically restricted to low-dimensional, tabular parameter spaces and cannot be directly used in more complex simulation setups. In contrast, many modern applications involve richer input configurations, including unstructured meshes, spatially varying fields, and complex boundary or material conditions. These settings require learning from heterogeneous, high-dimensional data with nonparametric variability.

Recent advances in physics machine learning have begun to address these challenges. One line of work, often referred to as physics-based model reduction, builds surrogates that approximate the solution of the governing equations directly~\cite{amsallem2015design, ijnme_casenave_2020, daniel2020model, lee2020model, KIM2022110841, barral2024registration}.
Other approaches have also been proposed using non-parametric methods based on the use of morphing~\cite{casenave2024mmgp, KABALAN2025115929, kabalan2025ommgp} or optimal transport~\cite{perez2024gaussianprocessregressionsliced, perez2025learningsignalsdefinedgraphs}, and have the advantage of requiring a smaller number of design points. Deep learning methods have recently shown strong potential for modeling the spatiotemporal dynamics of physical systems. Graph Neural Networks (GNNs) build on the message-passing paradigm introduced by~\citet{gilmer2017neural}: architectures such as MeshGraphNets~\cite{pfaff2021learning} extend GNNs to general mesh-based simulations. Hierarchical versions like MultiScale MeshGraphNets~\cite{fortunato2022multiscale} enhance scalability and accuracy, while recent works demonstrate effectiveness in inverse~\cite{allen2022physical} and steady-state problems~\cite{harsch2021direct}. Other developments include geodesic convolutions~\cite{baque2018geodesic}, multi-resolution models~\cite{lino2021simulating,lino2022multi}, and improved pooling strategies~\cite{cao2023efficient}. More recently, transformer-based architectures have been investigated for physics-based learning, leveraging attention mechanisms to capture long-range dependencies and complex interactions across mesh-based representations~\cite{dosovitskiy2020image, jiang2023transcfd, zhou2026transolver}. Tools such as PhysicsNeMo~\cite{physicsnemo2023}, PyTorch Geometric~\cite{pyg}, and Deep Graph Library~\cite{dgl} provide convenient foundations for implementing these methods.

Despite these advances, widespread adoption remains hindered by a critical bottleneck: the lack of large-scale, diverse, and standardized datasets for training and benchmarking. Existing datasets often cover narrow physical regimes, rely on ad hoc formats, or are tied to specific libraries, limiting reusability and interoperability. Furthermore, many datasets are tailored to isolated challenges (e.g., time dependence) but fail to accommodate others (e.g., geometric variation). This fragmentation is particularly problematic in the context of recent developments in physics foundation models~\cite{Yang_2023, mccabe2024multi, NEURIPS2024_d7cb9db5, Birk_2024, PhysRevE.111.035304}, which require large, flexible, and standardized sources of training data.


\textbf{Contributions.} To address these limitations, we present PLAID, a standardized data layer for physics simulation data that bridges simulation and machine learning workflows. It introduces a unified data model for heterogeneous mesh-based simulations, enabling interoperable datasets and a reproducible benchmark spanning multiple sources of heterogeneity:
\begin{itemize}
\item \textbf{Problem identification.} We identify heterogeneous simulation data (variable meshes, geometries, and topologies) as a key bottleneck in physics machine learning, limiting reliable evaluation and generalization.
\item \textbf{Unified data model.} We introduce PLAID (Physics-Learning AI Data model), a generic, flexible, and extensible representation for physics simulations, supporting time-dependent problems, remeshing, topology variability, and mixed-element unstructured meshes.
\item \textbf{Realistic dataset collection.} We release six datasets in structural mechanics and computational fluid dynamics, reflecting industrial scenarios with complex geometries and multi-source heterogeneity.
\item \textbf{Reproducible benchmarking.} We provide a standardized evaluation protocol and benchmark results across multiple modeling paradigms (GNNs, operator learning, transformers), enabling fair comparison under heterogeneous conditions.
\item \textbf{Open evaluation platform.} We extend the benchmark into a community-driven framework through online competitions hosted on Hugging Face, fostering continuous and scalable evaluation.
\end{itemize}
We provide an accompanying software library to facilitate dataset creation, reading, and high-level interaction, leveraging Hugging Face infrastructure for efficient streaming and sharing.


In Section~\ref{sec:related_work}, we review relevant dataset efforts in the literature. Section~\ref{sec:plaid} introduces the PLAID data model and implementation, along with six publicly released datasets in structural mechanics and computational fluid dynamics, presented in Section~\ref{sec:datasets}, that showcase rich variability in physics and numerical complexity. In Section~\ref{sec:benchmark}, we provide performance benchmarks across a range of machine learning methods, hosted on Hugging Face to allow community participation and continual updates. We conclude with perspectives in Section~\ref{sec:conclusion}.

\section{Related Work}
\label{sec:related_work}

Progress in machine learning has been largely driven by the availability of large, diverse, and carefully curated datasets~\cite{achiam2023gpt, touvron2023llamaopenefficientfoundation, meta2025llama4}. Natural language processing models are trained on web-scale data~\cite{gao2020pile, penedo2023refinedweb, LLMDataHub, liu2024datasetslargelanguagemodels}, and vision models routinely leverage billions of image–text pairs~\cite{schuhmann2022laion, bai2023, chen2024panda70m}.

In contrast, datasets for physics learning remain comparatively underdeveloped. Early benchmarks targeted standard physics problems and reference simulations~\cite{pdebench, pdearena, hao2023pinnacle}. More recent datasets have focused on complex, domain-specific settings~\cite{hersbach2020era5, bonnet2022airfrans, kohl2023_acdm,janny2023eagle, hassan2023bubbleml, chung2023turbulence_blastnet, yu2023climsim, toshev2024lagrangebench, ashton2024drivaerml}.
The Well, introduced by \citet{ohana2024well}, includes a large list of datasets for various physics, but is limited to structured grids (uniformly sampled domains).

Structural mechanics simulations, with non-linear constitutive laws, are of paramount importance for industrial design, and are poorly represented in available datasets. Most available datasets use data models that limit their evolution and generality. Complex industrial settings include vertices and element tags, heterogeneous data with remeshing, multiple meshes of various dimensions, topologies and mixed element types, compatible with commercial codes routinely used by design engineers. Besides, most datasets come with a library dedicated to the dataset, featuring specific commands and assumptions, which limit their wide adoption.

\textbf{Comparison with existing datasets.} Our goal is to support fair and reproducible surrogate modeling across heterogeneous simulation settings. Existing datasets are often limited in their ability to capture realistic industrial scenarios, including complex geometries, variable topologies, and heterogeneous data structures. To position PLAID within this landscape, we provide a comparison in Table~\ref{tab:dataset_comparison}, where “sample heterogeneity” refers to possible presence of multiple meshes per sample, with differing dimensions, topologies, element types and variable topology.

\begin{table*}[h!]
\small
\centering
\caption{Comparison of dataset collections across key characteristics.
${}^{*}$ Geometrical variations addressed via density.
“Physics” describes the physical families covered by the released datasets. PLAID currently covers structural mechanics and CFD, while the data model itself is designed to support broader physics families.}
\label{tab:dataset_comparison}
\begin{tabularx}{\textwidth}{L{5.38cm} C{0.73cm} C{1.13cm} C{0.93cm} C{1.08cm} C{1.33cm} C{1.58cm} C{1.3cm}}
  \toprule
  \texttt{Dataset collection} & Steady & Transient & 2D+3D & Complex domains & Geometrical variations & Sample hete-rogeneity & Physics \\
  \midrule
  \texttt{Mech.-MNIST}~\cite{lejeune2020mechanical} & \checkmark & \checkmark & $\times$ & $\times$ & $\times$ & $\times$ & Solids \\
  \texttt{PDEArena}~\cite{gupta2022towards} & $\times$ & \checkmark & \checkmark & $\times$ & $\times$ & $\times$ & Fluids \\
  \texttt{BubbleML}~\cite{hassan2023bubbleml} & $\times$ & \checkmark & \checkmark & $\times$ & $\times$ & $\times$ & Boiling \\
  \texttt{BLASTNet}~\cite{chung2023turbulence} & \checkmark & \checkmark & \checkmark & $\times$ & $\times$ & $\times$ & Fluids \\
  \texttt{PDEBench}~\cite{takamoto2022pdebench} & \checkmark & \checkmark & \checkmark & $\times$ & $\times$ & $\times$ & Fluids \\
  \texttt{The Well}~\cite{ohana2024well} & $\times$ & \checkmark & \checkmark & \checkmark & $\sim^{*}$ & $\times$ & Mixed \\
  \texttt{PINNacle}~\cite{hao2024pinnacle} & \checkmark & \checkmark & \checkmark & \checkmark & $\times$ & $\times$ & Mixed \\
  \texttt{PLAID (ours)} & \checkmark & \checkmark & \checkmark & \checkmark & \checkmark & \checkmark & Solids+Fluids\\
  \bottomrule
\end{tabularx}
\end{table*}

\section{PLAID standard}
\label{sec:plaid}


PLAID should be understood as three complementary layers: (i) a machine-learning-oriented data model for heterogeneous physics simulations, (ii) an accompanying open-source library implementing this model through high-level accessors and dataset construction utilities~\href{github.com/PLAID-lib/plaid}{GitHub}~\cite{plaid2025}, and (iii) a benchmark and dataset-distribution layer integrated with modern ML infrastructure.

\paragraph{Relation to CGNS~\cite{poinot2018seven}.} PLAID does not aim to replace CGNS. Instead, it uses CGNS-compatible hierarchical simulation trees as one of its core storage/schema representations and adds ML-specific conventions on top of them: problem definitions, sample-level metadata, input/output field and scalar declarations, train/test split metadata, time-indexed accessors, backend serialization, and standardized benchmark/evaluation interfaces. In this sense, CGNS standardizes simulation data structures, whereas PLAID standardizes reusable machine-learning datasets and benchmark tasks derived from heterogeneous simulations.

\begin{figure}[h!]
    \centering
    \includegraphics[width=\linewidth]{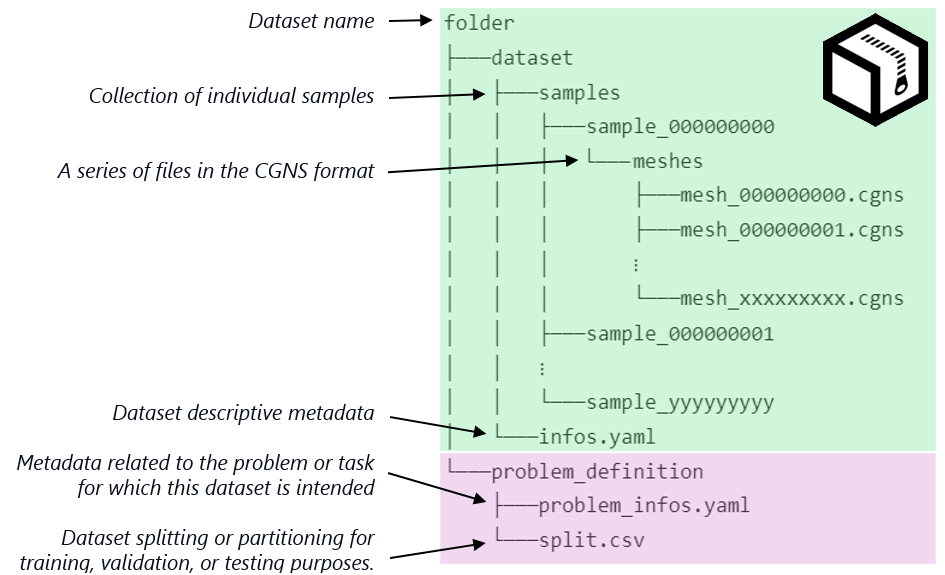}
    \caption{PLAID files structure (CGNS backend).}
    \label{fig:plaid_format}
\end{figure}

PLAID datasets are provided either in a human-readable format or through efficient storage backends such as Hugging Face Datasets~\cite{hf_datasets} and Zarr~\cite{zarr}. In the former case, YAML files can be opened with any text editor, while CGNS files containing physical configurations can be visualized using tools such as ParaView, see Figure~\ref{fig:plaid_format}. In the latter case, these backends provide advanced data management capabilities, including online streaming and feature-wise access, allowing subsets of variables to be read without loading the entire dataset.

Additionally, PLAID offers high-level utilities for constructing, handling and parallel reading/writing datasets efficiently. Documentation is available \href{plaid-lib.readthedocs.io}{online},
with usage examples and tutorials showing how users can create a PLAID dataset from their own data. We also mention Muscat, a finite element toolbox available on \href{gitlab.com/drti/muscat}{GitLab}~\cite{Bordeu2023, muscat2023}, containing various readers and writers for simulation file formats used in numerical simulation codes for physics, and routines to generate the CGNS data structures used in PLAID. Samples can feature multiple meshes, scalars, fields and time series. To illustrate how PLAID handles such heterogeneous data, we highlight a few representative accessors:
\begin{itemize}
\item {\small{\texttt{sample.get\_field\_names(name, zone\_name, base\_name, location, time)}}}: retrieves a field identified by {\small{\texttt{name}}} within the specified {\small{\texttt{zone\_name}}}, {\small{\texttt{base\_name}}}, {\small{\texttt{location}}} (e.g., Vertex, CellCenter, FaceCenter), and {\small{\texttt{time}}} in the CGNS structure. Both fields and meshes may evolve over time, enabling support for remeshing and dynamic field appearance or disappearance.
\item {\small{\texttt{sample.get\_field(name)}}}: provides simplified access by automatically handling default values, avoiding the need to specify {\small{\texttt{zone\_name}}}, {\small{\texttt{base\_name}}}, and {\small{\texttt{location}}} in unambiguous cases.
\item {\small{\texttt{sample.show\_tree(time)}}}: returns a summary of the CGNS tree at the specified time step.
\end{itemize}
The same mechanism naturally extends to samples composed of several geometrical supports. For instance, \texttt{VKI-LS59} stores the two-dimensional fluid domain and the one-dimensional blade-surface domain in separate CGNS bases, with fields attached to the appropriate support. This multi-base structure is the intended representation for coupled or multi-physics simulations where different quantities live on different meshes. More examples are provided in Appendix~\ref{app:plaid}.

\section{PLAID datasets}
\label{sec:datasets}

Among the datasets presented below, some (\texttt{Tensile2D}, \texttt{2D\_MultiScHypEl}, \texttt{Rotor37} and \texttt{AirfRANS}) have been previously introduced in the literature, while others (\texttt{2D\_ElPlDynamics}, \texttt{2D\_profile}, and \texttt{VKI-LS59}) are newly released here. All are unified within the PLAID standard.

\subsection{Dataset design rationale.}

The datasets are designed to span increasing levels of complexity and realism, from controlled settings to industrial-scale simulations involving heterogeneous geometries, variable topology, and nonlinear physics. Table~\ref{tab:plaid_datasets} summarizes the motivation behind each dataset in terms of machine learning challenges and industrial relevance.

All datasets involve variable geometries, a key aspect of industrial design, and are generated using simulation codes employed in industry (Z-set, OpenRadioss, elsA, and BROADCAST), together with constitutive laws for solids and turbulence models used in practice. For instance, \texttt{Rotor37} corresponds to the surface solution of a full 3D CFD simulation with complex physics. \texttt{VKI-LS59} reflects production-level CFD simulations and involves multiple geometrical supports of different dimensionality, while \texttt{2D\_ElPlDynamics} introduces time-dependent behavior with highly nonlinear phenomena such as damage, fracture, and non-local constitutive laws (see~\ref{app:plaid-vki} and~\ref{app:plaid-plastodyn}). This progression in complexity is essential to evaluate machine learning models beyond simplified settings and highlights the challenges posed by heterogeneous data in realistic scenarios.

\begin{table*}[h!]
\footnotesize
\centering
\caption{Overview of PLAID datasets, associated machine learning challenges, and industrial relevance.}
\begin{tabularx}{\textwidth}{L{2.6cm} L{6.7cm} L{6.7cm}}
  \toprule
  \texttt{Dataset} & ML Challenge & Industrial relevance \\
  \midrule
  \texttt{Tensile2d} & {\footnotesize Unstruct. mesh, variable size fields, nonlinear laws} & {\footnotesize Metallic materials with elastoviscoplastic behavior} \\
  \texttt{2D\_MultiScHypEl} & {\footnotesize Unstruct. mesh, variable size fields, variable topology} & Representative volume elements for bi-materials \\
  \texttt{2D\_ElPlDynamics} & Time-dependent, topology variation, nonlin. materials & Structural integrity in extreme conditions \\
  \texttt{Rotor37} & 3D, shocks of variable position, nonlinear model & Design of rotors in compressors of rotating machinery \\
  \texttt{2D\_profile} & Variable size fields, variable shock positions, nonlin. & Design of wings and propellers \\
  \texttt{VKI-LS59} & Variable shock classes, periodic setting, nonlin. & Design of rotors in turbines of rotating machinery \\
  \bottomrule
\end{tabularx}
\label{tab:plaid_datasets}
\end{table*}

\subsection{Structural mechanics}

\subsubsection{\texttt{Tensile2d}~\cite{casenave_2025_14840177} (\href{https://zenodo.org/records/14840177}{Zenodo}, \href{https://huggingface.co/datasets/PLAID-datasets/Tensile2d}{Hugging Face})}

\texttt{Tensile2d} is a simple dataset of 2D quasistatic non-linear structural mechanics simulations, in small deformations and plane strain regimes, solved with Z-set~\cite{zset} using the finite element method. The material is modeled with a non-linear constitutive law. The dataset computes the deformation of a structure subjected to an imposed negative constant pressure at the top, and zero displacement at the bottom, see Figure~\ref{fig:tensile2d_emmental_setting} (left). Only the steady-state solution is kept.

Input variability in the dataset are the unstructured meshes (variable shape, number of nodes and connectivity), the pressure P at the top boundary condition (scalar) and 5 scalars modeling the non-linear constitutive law: (\texttt{p1}, \texttt{p2}, \texttt{p3}, \texttt{p4} and \texttt{p5}). Outputs of interest are 4 scalars (\texttt{max\_von\_mises}, \texttt{max\_q}, \texttt{max\_U2\_top} and \texttt{max\_sig22\_top}) and 6 fields (\texttt{U1}, \texttt{U2}, \texttt{q}, \texttt{sig11}, \texttt{sig22} and \texttt{sig12}). Seven nested training sets are provided, as well as a testing set and two out-of-distribution samples.

\subsubsection{\texttt{2D\_MultiScHypEl}~\cite{staber_2025_14840446} (\href{https://zenodo.org/records/14840446}{Zenodo}, \href{https://huggingface.co/datasets/PLAID-datasets/2D_Multiscale_Hyperelasticity}{Hugging Face})}

\begin{figure}[h!]
    \centering
    \includegraphics[width=0.56\linewidth]{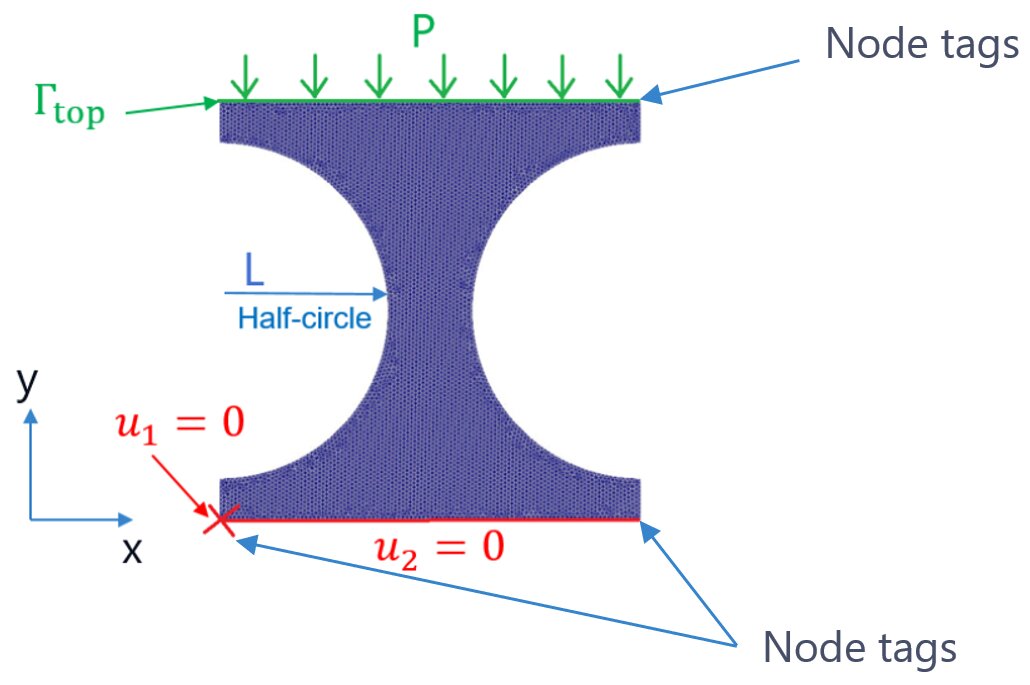}
    \includegraphics[width=0.40\linewidth]{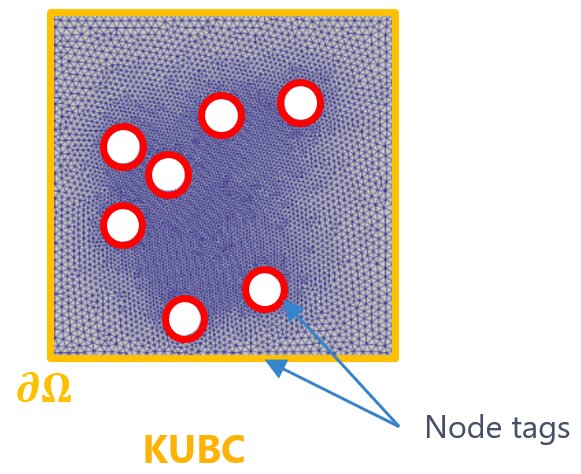}
    \caption{\texttt{Tensile2d} (left) and \texttt{2D\_MultiScHypEl} (right).}
    \label{fig:tensile2d_emmental_setting}
\end{figure}

\texttt{2D\_MultiScHypEl}, standing for 2D multiscale hyperelasticity, is a dataset of 2D quasistatic non-linear structural mechanics simulations under large deformation and plane strain conditions, solved with DOLFINx~\cite{dolfinx} using the finite element method. The material behavior follows a compressible hyperelastic constitutive law, capturing complex non-linear responses. Each simulation corresponds to the homogenization of a porous representative volume element (RVE), subject to kinematically uniform boundary conditions (KUBC)~\cite{yvonnet2019computational}, see Figure~\ref{fig:tensile2d_emmental_setting} (right).

Input variability in the dataset are the unstructured meshes (variable shape, number of nodes, connectivity and topology--the number of circular inclusions) and the 3 scalars modeling the KUBC, namely the components \texttt{C11}, \texttt{C12}, and \texttt{C22} of the macroscopic right Cauchy-Green deformation tensor. Outputs of interest are 1 scalar (\texttt{effective\_energy}) and 7 fields (displacements \texttt{u1}, \texttt{u2}; first Piola-Kirchhoff stress components \texttt{P11}, \texttt{P12}, \texttt{P22}, \texttt{P21} and the strain energy density field \texttt{psi}). Various training and testing sets are provided (both across all topologies and within each topology~class).

\subsubsection{\texttt{2D\_ElPlDynamics}~\cite{piat_2025_15286369} (\href{https://zenodo.org/records/15286369}{Zenodo}, \href{https://huggingface.co/datasets/PLAID-datasets/2D_ElastoPlastoDynamics}{Hugging Face})}

\texttt{2D\_ElPlDynamics}, standing for 2D elasto-plasto dynamics, is a dataset of 2D dynamic non-linear structural mechanics simulations, in large deformations and plane strain regimes, solved with OpenRadioss~\cite{openradioss} using the finite element method. The material is modeled with a non-linear elastoplastic law, with damage (modeled using element erosion), failure and a non-local method for reducing mesh sensitivity. The dataset computes the transient deformation of a 2D structure, subjected to imposed displacement on the right and zero displacement on the left, see Figure~\ref{fig:epdyn_rotor37} (left).

Input variability in the dataset are the unstructured meshes (variable shape, number of nodes, connectivity and topology). Outputs of interest are 3 fields (\texttt{U\_x} and \texttt{U\_y} the displacement fields at the nodes, and \texttt{EROSION\_STATUS} a boolean field at the elements describing the state -- valid or broken -- of each element). A training and a testing set are provided.

\subsection{Computational fluid dynamics}

\subsubsection{\texttt{Rotor37}~\cite{roynard_2025_14840190} (\href{https://zenodo.org/records/14840190}{Zenodo}, \href{https://huggingface.co/datasets/PLAID-datasets/Rotor37}{Hugging Face})}

\begin{figure}[h!]
    \centering
    \includegraphics[width=0.58\linewidth]{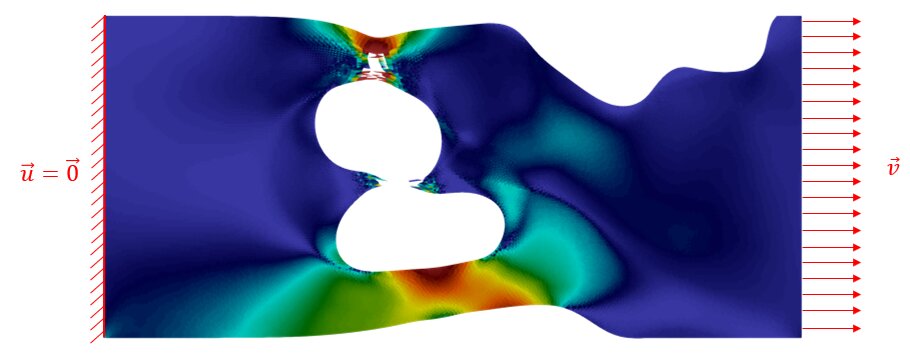}
    \includegraphics[width=0.40\linewidth]{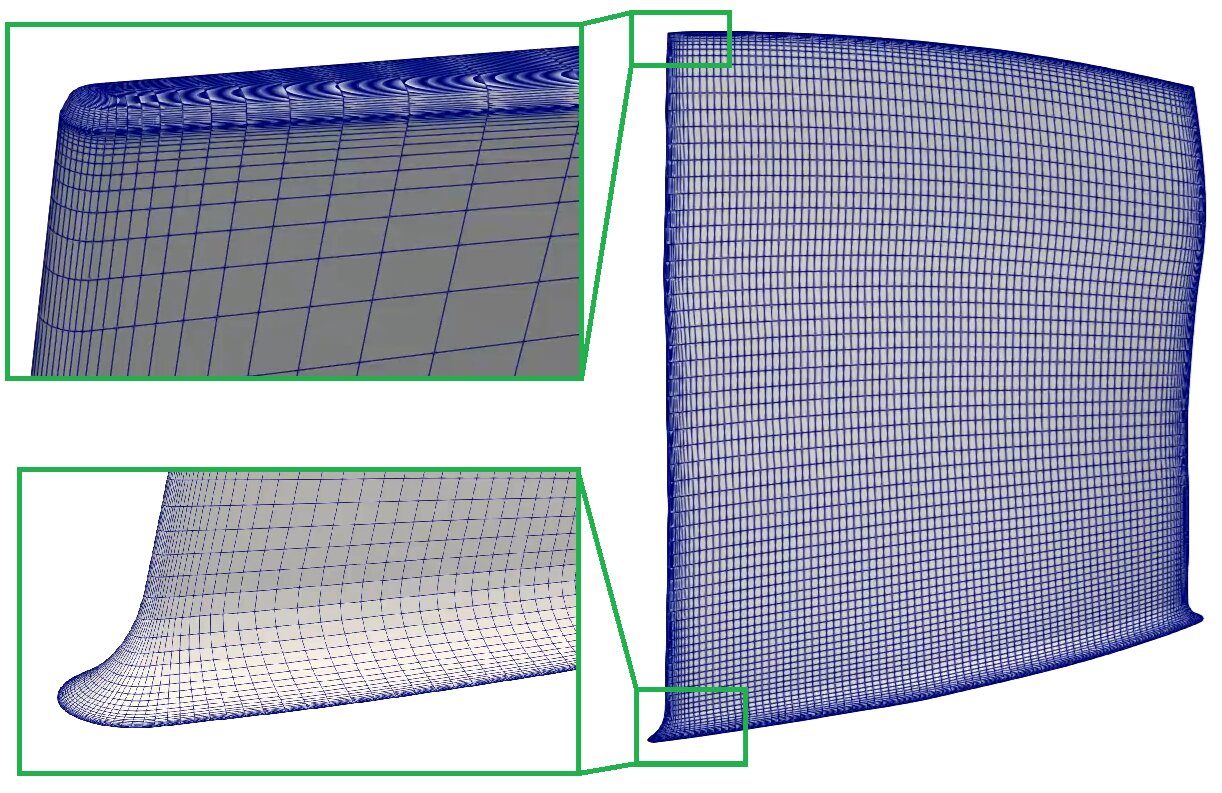}
    \caption{\texttt{2D\_ElPlDynamics} (left) and \texttt{Rotor37} (right).}
    \label{fig:epdyn_rotor37}
\end{figure}

\texttt{Rotor37} is a dataset of 3D compressible steady-state Reynolds-Averaged Navier-Stokes (RANS) simulations, solved with elsA~\cite{elsa} using the finite volume method. Large scale simulations around the rotor37 blade inside a 3D duct have been computed, with inflow, outflow and periodic boundary conditions. An adequate turbulence model and laws of the wall have been chosen. The dataset only keeps the steady-state solution at the boundary of the blade inside the duct, and scalars of interest, see Figure~\ref{fig:epdyn_rotor37} (right).

Input variability in the dataset are the block-structured anisotropic meshes (variable shape, normals at the blade surface are provided) and 2 scalars (the pressure \texttt{P} and the rotation speed \texttt{Omega} of the blade). Outputs of interest are 3 scalars (\texttt{Massflow}, \texttt{Compression\_ratio} and \texttt{Efficiency}) and 3 fields (\texttt{Density}, \texttt{Pressure}, \texttt{Temperature}). Eight nested training sets and a testing set are provided.

\subsubsection{\texttt{2D\_profile}~\cite{casenave_2025_15155119} (\href{https://zenodo.org/records/15155119}{Zenodo}, \href{https://huggingface.co/datasets/PLAID-datasets/2D_profile}{Hugging Face})}

\texttt{2D\_profile} is a dataset of 2D compressible steady-state Reynolds-Averaged Navier-Stokes (RANS) simulations, solved with elsA~\cite{elsa} using the finite volume method. The flow is computed around 2D profiles, which present large deformation around shapes resembling airfoils or propeller blades, on large refined meshes, with inflow, outflow and periodic boundary conditions, at a transonic regime. An adequate turbulence model and laws of the wall have been chosen. The dataset only keeps the steady-state solution on a zone cropped close to the profile, see Figure~\ref{fig:profile_vki} (left).

Input variability in the dataset are the unstructured anisotropic meshes (variable shape, number of nodes and connectivity). Outputs of interest are 4 fields (\texttt{Mach}, \texttt{Pressure}, \texttt{Velocity-x} and \texttt{Velocity-y}). A training and a testing set are provided.

\subsubsection{\texttt{VKI-LS59}~\cite{bucci_2025_14840512} (\href{https://zenodo.org/records/14840512}{Zenodo}, \href{https://huggingface.co/datasets/PLAID-datasets/VKI-LS59}{Hugging Face})}

\begin{figure}[h!]
    \centering
    \includegraphics[width=0.24\linewidth]{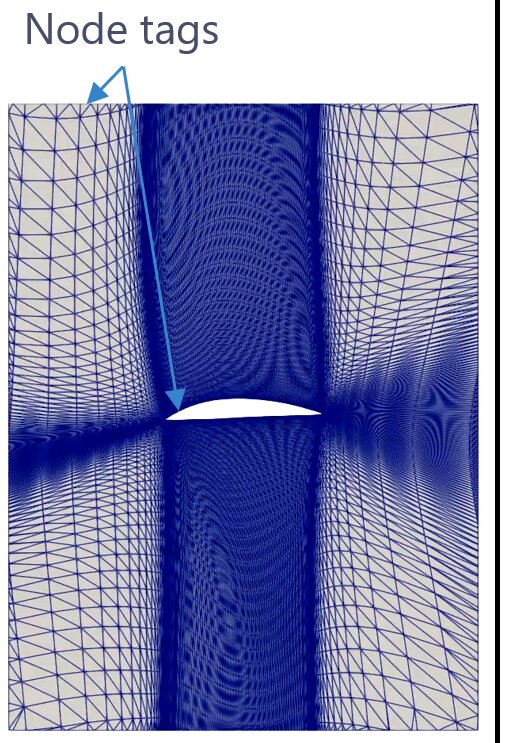}
    \includegraphics[width=0.74\linewidth]{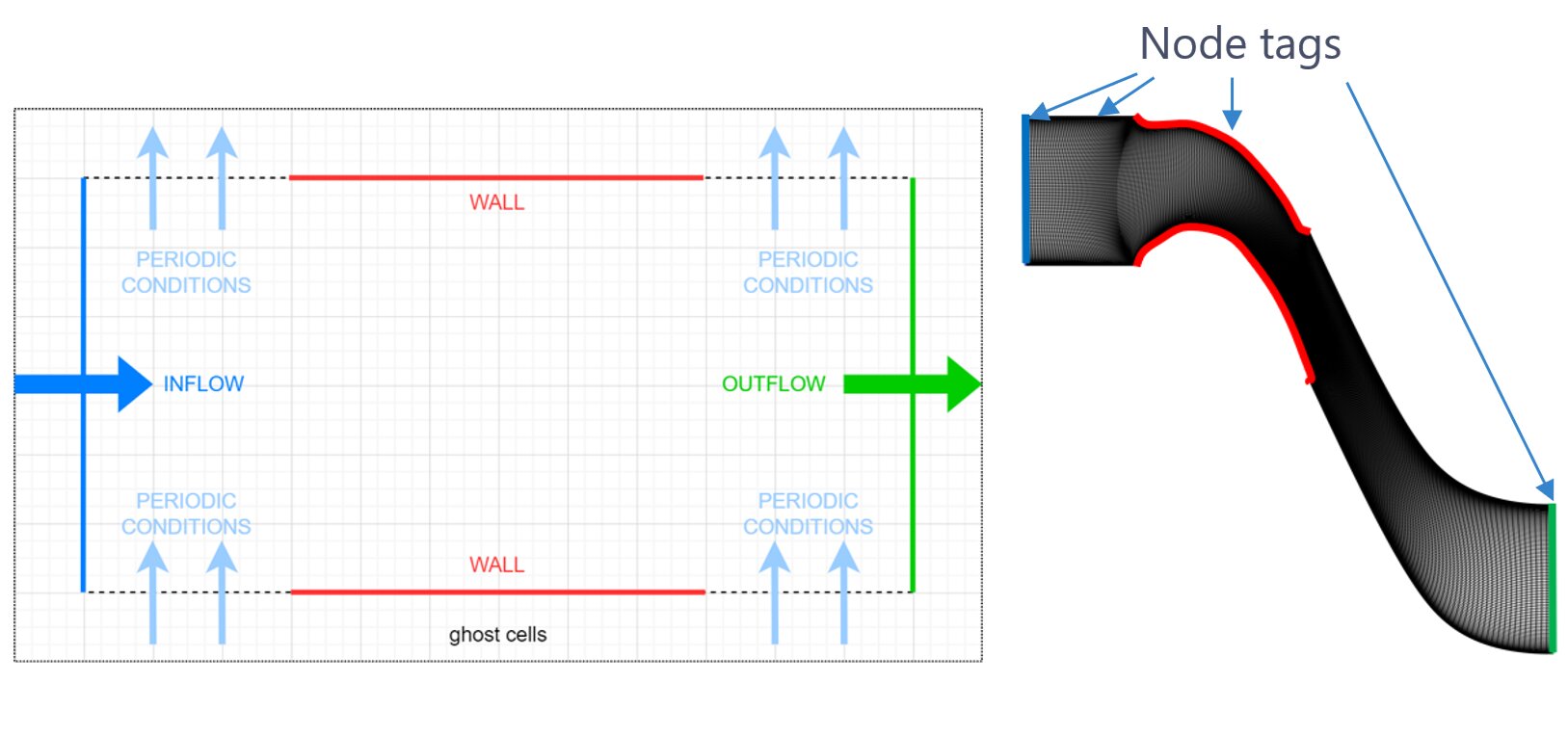}
    \caption{\texttt{2D\_profile} (left) and \texttt{VKI-LS59} (right).}
    \label{fig:profile_vki}
\end{figure}

\texttt{VKI-LS59} is a dataset of 2D compressible steady-state Reynolds-Averaged Navier-Stokes (RANS) simulations, solved with BROADCAST~\cite{poulain2023broadcast} using the finite volume method with high-order corrections. The flow is computed around the VKI-LS59 blade, with inflow, outflow and periodic boundary conditions. A Spalart-Allmaras turbulence model has been chosen, see Figure~\ref{fig:profile_vki} (right).

Input variability in the dataset are the block-structured anisotropic meshes (variable shape, number of nodes and connectivity, the distance field to the blade surface is provided) and 2 scalars (\texttt{angle\_in} and \texttt{mach\_out}). Outputs of interest are 6 scalars (\texttt{Q}, \texttt{power}, \texttt{Pr}, \texttt{Tr}, \texttt{eth\_is} and \texttt{angle\_out}) and 7 fields (\texttt{ro}, \texttt{rou}, \texttt{rov}, \texttt{roe}, \texttt{nut}, \texttt{mach} and \texttt{M\_iso} -- this last being only defined at the surface of the blade). Eight nested training sets are provided, as well as a testing set.

\subsubsection{\texttt{AirfRANS}~\cite{bonnet2022airfrans}}

\texttt{AirfRANS} is a dataset of external aerodynamics, featuring steady-state Reynolds-Averaged Navier-Stokes (RANS) simulations over airfoils at a subsonic regime, proposed by~\citet{bonnet2022airfrans}, which we refer to for a detailed description. In addition to the six original datasets, we provide three variants of \texttt{AirfRANS} in PLAID format: original~\cite{roynard_2025_14840387}(\href{https://zenodo.org/records/14840387}{Zenodo}, \href{https://huggingface.co/datasets/PLAID-datasets/AirfRANS_original}{Hugging Face}), clipped~\cite{roynard_2025_14840377}(\href{https://zenodo.org/records/14840377}{Zenodo}, \href{https://huggingface.co/datasets/PLAID-datasets/AirfRANS_clipped}{Hugging Face}) and remeshed~\cite{roynard_2025_14840388}(\href{https://zenodo.org/records/14840388}{Zenodo}, \href{https://huggingface.co/datasets/PLAID-datasets/AirfRANS_remeshed}{Hugging Face}).

Input variability in the dataset are the anisotropic meshes (variable shape, number of nodes and connectivity, the distance field to the airfoil surface is provided) and 2 scalars (\texttt{angle\_of\_attack} and \texttt{inlet\_velocity}). Outputs of interest are 2 scalars (\texttt{C\_D} and \texttt{C\_L}) and 4 fields (\texttt{nut}, \texttt{Ux}, \texttt{Uy} and \texttt{p}). Various training and testing sets are provided.

\subsection{Dataset collection}

The collection of proposed datasets is publicly available through dedicated~\href{https://zenodo.org/communities/plaid_datasets}{Zenodo} and~\href{https://huggingface.co/PLAID-datasets}{Hugging Face} communities.
It covers a diverse set of challenging scenarios, including heterogeneous meshes, variable topology, structural mechanics and CFD applications, and time-dependent problems. While the collection can be further extended, it already provides a practical basis for benchmarking surrogate models in realistic industrial settings.

Since these datasets are intended as open benchmarks for the community, outputs on the test sets are not released; instead, evaluation tools are provided to compute scores. We provide high-level descriptions of the underlying physical models and assumptions in Tables~\ref{tab:dataset_description_1} and~\ref{tab:dataset_description_2}, while avoiding full disclosure of simulation details that would enable reconstruction of the high-fidelity solvers and compromise the integrity of the benchmark. Some field outputs are illustrated in Table~\ref{fig:dataset_solutions}.

\begin{table*}[h!]
\small
\centering
  \caption{Dataset collection description: model and simulation volume.}
  \begin{tabularx}{\textwidth}{L{3.4cm}C{1.62cm}C{6cm}C{1.31cm}C{1.25cm}C{1.25cm}}
  \toprule
 \texttt{Dataset} & Simulation code & Model & Nb samples & Volume Zenodo & Volume HF \\
  \midrule
  \small{\texttt{Tensile2d}}  & Z-set & 2D quasistatic non-linear structural mechanics, small deformations, non-linear constitutive law & 702 & 290\phantom{.}MB & 383\phantom{.}MB\\
  \small{\texttt{2D\_MultiScHypEl}} & DOLFINx & 2D quasistatic non-linear structural mechanics, finite elasticity & 1,140 & 350\phantom{.}MB & 419\phantom{.}MB\\
  \small{\texttt{2D\_ElPlDynamics}} & OpenRadioss & 2D dynamic non-linear structural mechanics, non-linear non-local constitutive law & 1,018 & 5.7\phantom{.}GB & 8.6\phantom{.}GB\\
  \small{\texttt{Rotor37}}  & elsA & 3D Navier-Stokes (RANS) & 1,200 & 3.3\phantom{.}GB & 4.0\phantom{.}GB\\
  \small{\texttt{2D\_profile}} & elsA & 2D Navier-Stokes (RANS) & 400 & 660\phantom{.}MB & 814\phantom{.}MB\\
  \small{\texttt{VKI-LS59}} & BROADCAST & 2D Navier-Stokes (RANS)& 839 & 1.9\phantom{.}GB & 2.3\phantom{.}GB\\
  \midrule
  \small{\texttt{AirfRANS original}}  &  &  &  & 9.3\phantom{.}GB & 15.6\phantom{.}GB\\
  \small{\texttt{AirfRANS clipped}}  & OpenFOAM  & 2D Navier-Stokes (RANS) & 1,000 & 9.7\phantom{.}GB & 18.2\phantom{.}GB\\
  \small{\texttt{AirfRANS remeshed}}  &  &  &  & 520\phantom{.}MB & 611\phantom{.}MB\\
  \bottomrule
 \end{tabularx}
 \label{tab:dataset_description_1}
\end{table*}

\begin{table*}[h!]
\small
\centering
  \caption{Dataset collection description: data and splits, a $*$ in the second column means that the number of nodes and connectivity are constant in the dataset -- the position of the nodes still varies.}
  \begin{tabularx}{\textwidth}{L{3.62cm} L{3.25cm} L{2.5cm} L{2.87cm} L{3.5cm}}
  \toprule
 \texttt{Dataset} & Mesh (mean nodes) & Inputs & Outputs & Splits (train/test)\\
  \midrule
  \small{\texttt{Tensile2d}}  & tri (9,428) & mesh, 6 scalars & 4 scalars, 6 fields & 500 / 200\\
  \small{\texttt{2D\_MultiScHypEl}} & tri (5,692) & mesh, 3 scalars & 1 scalar, 7 fields & 764 / 376 \\
  \small{\texttt{2D\_ElPlDynamics}} & tri (25,429) & mesh & 3 fields & 1,000 / 18 \\
  \small{\texttt{Rotor37}}  & quad (29,773*) & mesh, 2 scalars & 4 scalars, 3 fields & 1,000 / 200 \\
  \small{\texttt{2D\_profile}} & tri (37,042) & mesh & 4 fields & 300 / 100 \\
  \small{\texttt{VKI-LS59}} & quad (36,421*) & mesh, 2 scalars & 6 scalars, 7 fields & 671 / 168\\
  \midrule
  \small{\texttt{AirfRANS original}}  & quad (179,776) &    &  & \\
  \small{\texttt{AirfRANS clipped}}  & tri (179,779) & mesh, 2 fields & 2 scalars, 4 fields & various splits\\
  \small{\texttt{AirfRANS remeshed}}  & tri (7,624)  &  &  & \\
  \bottomrule
 \end{tabularx}
 \label{tab:dataset_description_2}
\end{table*}

\newlength{\mywidth}
\setlength{\mywidth}{0.8\textwidth}
\begin{table*}[h!]
\caption{Dataset collection examples of field outputs illustrations.}
\begin{tblr}{
  colspec = {Q[c]Q[c]},
  rowspec={|Q[m]|Q[m]Q[m]Q[m]Q[m]Q[m]Q[m]|},
  stretch = 0,
  rowsep = 3pt,
}
        \texttt{Dataset} & Examples of field outputs\\
        \small{\texttt{Tensile2d}} & \includegraphics[width=\mywidth, valign=c]{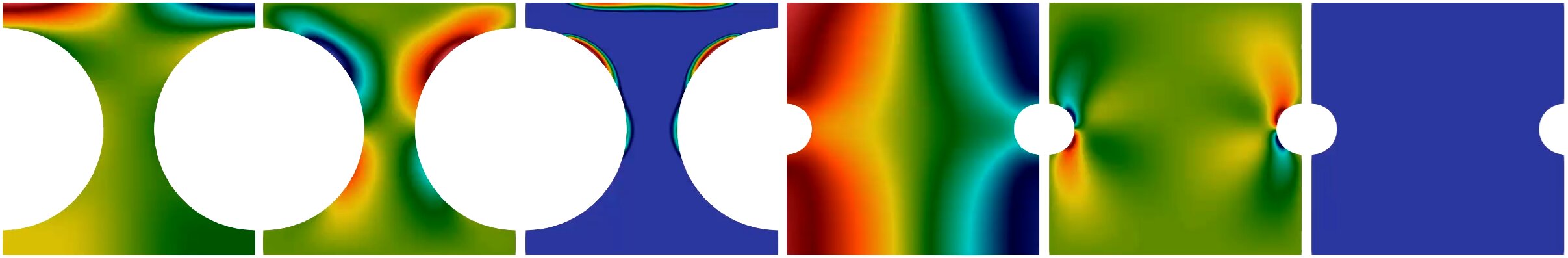} \\
        \small{\texttt{2D\_MultiScHypEl}} & \includegraphics[width=\mywidth, valign=c]{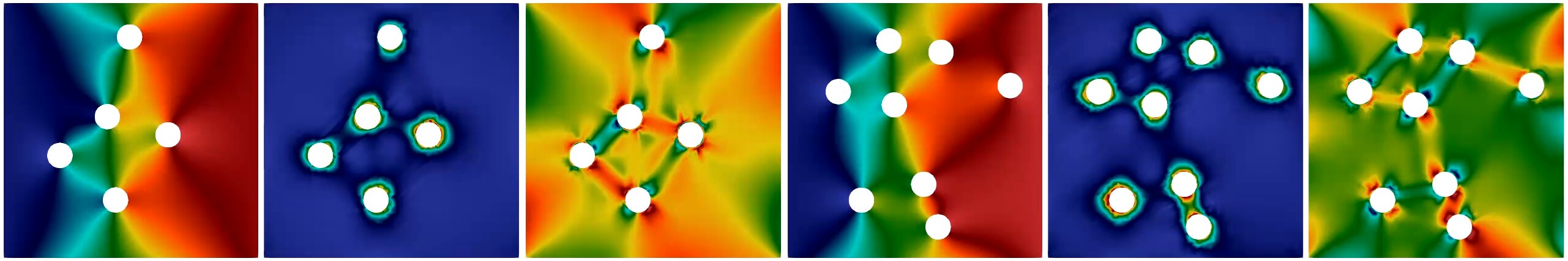} \\
        \small{\texttt{2D\_ElPlDynamics}} & \includegraphics[width=\mywidth, valign=c]{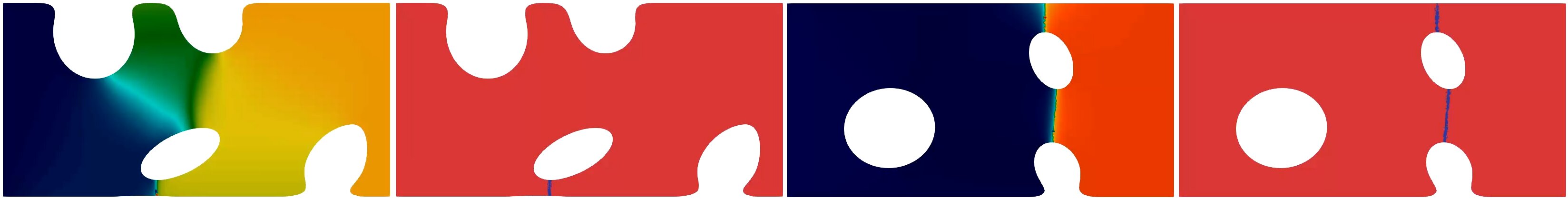} \\
        \small{\texttt{Rotor37}} & \includegraphics[width=\mywidth, valign=c]{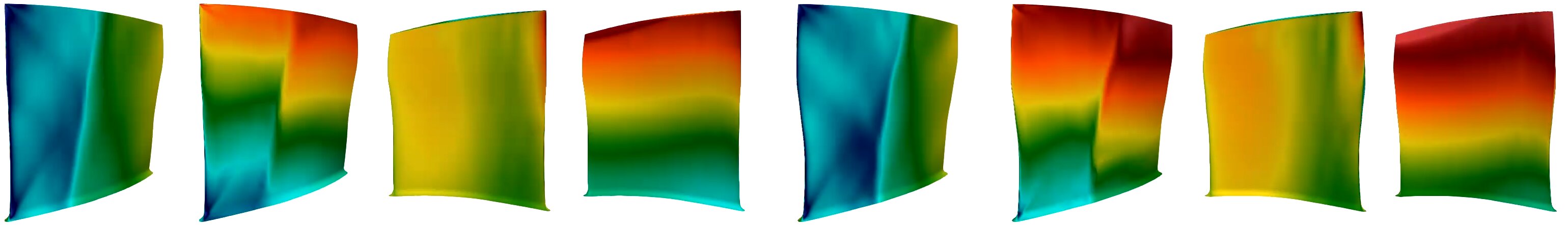} \\
        \small{\texttt{2D\_profile}} & \includegraphics[width=\mywidth, valign=c]{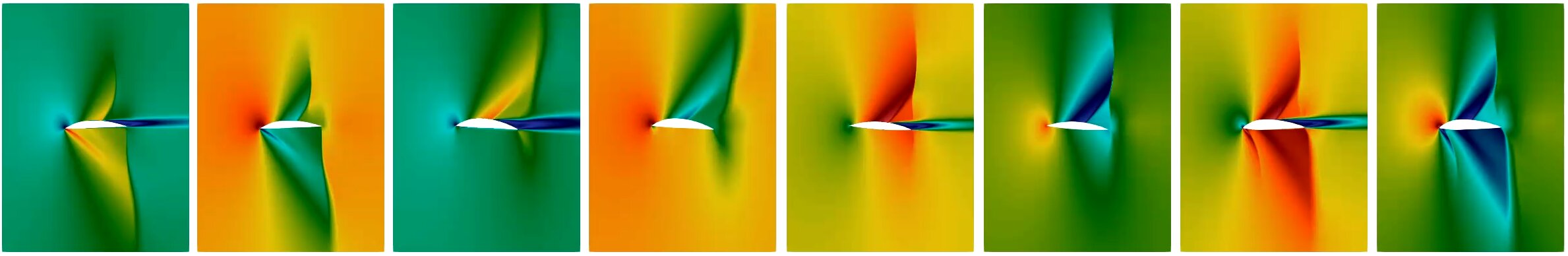} \\
        \small{\texttt{VKI-LS59}} & \includegraphics[width=\mywidth, valign=c]{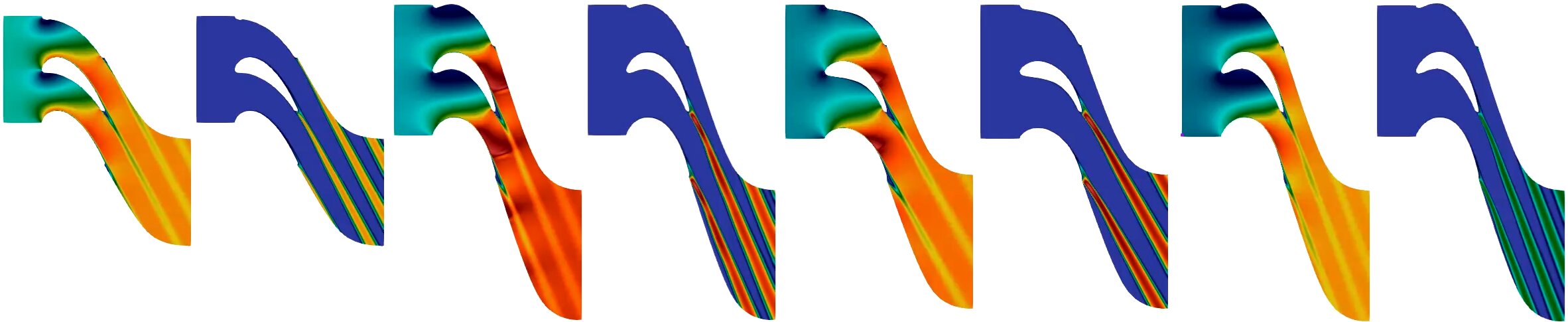} \\
\end{tblr}
    \label{fig:dataset_solutions}
\end{table*}

\section{Benchmark}
\label{sec:benchmark}

We mention that we do not provide benchmark tools and results for the \texttt{AirfRANS} datasets, since outputs are public on the testing sets, and various benchmarks have already been reported by~\citet{casenave2024mmgp} and~\citet{bonnet2022airfrans}, as well as in a competition at NeurIPS 2024~\citep{yagoubi_ml4cfd}.

\subsection{Methods}

We focus the benchmark on a representative set of widely used methods, covering graph-based, operator-learning, and transformer-based approaches, in order to provide a meaningful comparison across distinct modeling paradigms:
\begin{itemize}
\item MeshGraphNets (MGNs)~\cite{pfaff2021learning} are graph neural networks that utilize an encode-process-decode architecture, transforming mesh data into graph structures, processing them through message passing, and decoding the results to predict field outputs.
\item Mesh Morphing Gaussian Processes (MMGP)~\cite{casenave2024mmgp} rely on mesh morphing, finite element interpolation and dimensionality reduction to pretreat mesh-based data into a low dimensional embedding, and utilizes Gaussian processes to predict output scalars and fields.
\item Vi-Transformer~\cite{dosovitskiy2021image} and Augur~\footnote[1]{\href{https://augurco.fr/}{Augur} is a commercial solution.} rely on mesh partitioning to build tokens related to local mesh information and utilize a transformer to predict scalar and field outputs.
\item Fourier Neural Operators (FNOs)~\cite{li2020fourier} learn mappings between functions by operating in the Fourier domain.
\item Modulated Aerodynamic Resolution Invariant Operator (MARIO), introduced by~\citet{mario_paper}, builds upon~\citet{catalani2024neural} and exploits implicit neural representations, which model continuous signals by mapping input coordinates directly to output values, without relying on discrete grids or explicit storage.
\end{itemize}

For more details on the methods and their respective advantages/drawbacks, refer to Appendix~\ref{app:mldetails}.

\subsection{Evaluation metric}
\label{sec:eval_metric}

Accuracy of the trained models is evaluated by computing RRMSEs (Relative Root Mean Square Errors). Let $\{ \mathbf{f}^i_{\rm ref} \}_{i=1}^{n_\star}$ and $\{\mathbf{f}^i_{\rm pred} \}_{i=1}^{n_\star}$ be respectively the reference and prediction of a field output on the testing set. The RRMSE is defined as
\begin{align*}
    \mathrm{RRMSE}_f(\mathbf{f}_{\rm ref}, \mathbf{f}_{\rm pred}) = \left( \frac{1}{n_\star}\sum_{i=1}^{n_\star} \frac{\frac{1}{N^i}\|\mathbf{f}^i_{\rm ref} - \mathbf{f}^i_{\rm pred}\|_2^2}{\|\mathbf{f}^i_{\rm ref}\|_\infty^2} \right)^{1/2}\,,
\end{align*}
where $N^i$ is the number of nodes in the mesh of sample $i$, $n_\star$ is the number of samples in the testing set, and $\|\mathbf{f}^i_{\rm ref}\|_\infty$ is the maximum component in the vector $\mathbf{f}^i_{\rm ref}$. Similarly for scalar outputs, the following relative RMSE is computed:
\begin{align*}
    \mathrm{RRMSE}_s(\mathbf{s}_{\rm ref}, \mathbf{s}_{\rm pred}) = \left( \frac{1}{n_\star} \sum_{i=1}^{n_\star} \frac{|s^i_{\rm ref} - s_{\rm pred}^i|^2}{|s^i_{\rm ref}|^2} \right)^{1/2}\,.
\end{align*}
The score of a submission, \texttt{total\_error}, is the mean over fields and scalars RRMSEs. For time-dependent datasets, the RRMSE is computed over the full predicted trajectory by stacking field values over all available time steps. Thus, \texttt{2D\_ElPlDynamics} evaluates trajectory-level accuracy, while time-resolved diagnostics such as rollout-error growth curves are left for future work.

We mention that scalar quantities correspond to application-specific performance indicators (e.g., efficiency, stress extrema), ensuring that the evaluation reflects domain-relevant criteria.

\subsection{Benchmark results}
\label{sec:bench_res}

The global \texttt{total\_error} for each method and dataset is reported in Table~\ref{tab:PLAID_benchmark_short}, while all individual RRMSE metrics are provided in Table~\ref{tab:PLAID_benchmark} in Annex~\ref{app:benchdetails}.
These results define a fixed and fully reproducible reference benchmark, based on standardized datasets, fixed data splits, evaluation protocols, and implementations publicly available at submission time.

\begin{table*}[h!]
\small
\centering
  \caption{total\_error on PLAID benchmarks, best on each line is {\bf bold}, second best is \underline{underlined}. Missing entries for MMGP correspond to non-applicable settings.}
 \begin{tabularx}{\textwidth}{L{3.12cm}C{1.8cm}C{1.8cm}C{1.8cm}C{1.8cm}C{1.8cm}C{1.8cm}}
 \toprule
 \bf{total\_error} & MGN  & MMGP & Vi-Transf. & Augur & FNO & MARIO \\
  \midrule
  {\texttt{Tensile2d}}  & 0.0673 & \bf{0.0026} & 0.0116 & 0.0154 & 0.0123 & \underline{0.0038}  \\
  {\texttt{2D\_MultiScHypEl}}  & 0.0437 & - & 0.0325 & \bf{0.0232} & \underline{0.0302} & 0.0573   \\
  {\texttt{2D\_ElPlDynamics}} & 0.1202 & - & \underline{0.0227} & 0.0346 &  \bf{0.0215} & 0.0319 \\
  {\texttt{Rotor37}}  & 0.0074 & \bf{0.0014} & 0.0029 & 0.0033 & 0.0313 & \underline{0.0017}  \\
  {\texttt{2D\_profile}}  & 0.0593 & 0.0365 & \underline{0.0312} & 0.0425 & 0.0972 & \bf{0.0307}  \\
  {\texttt{VKI-LS59}}  & 0.0684 & 0.0312 & \underline{0.0193} & 0.0267 & 0.0215 & \bf{0.0124} \\
  \bottomrule
 \end{tabularx}
 \label{tab:PLAID_benchmark_short}
\end{table*}


Beyond the fixed benchmark defined in this work, we introduce online benchmarking applications hosted on Hugging Face as competitions with no end date, see \href{https://huggingface.co/PLAIDcompetitions}{Hugging Face benchmark collection}.
These applications extend the benchmark into a long-term, community-driven evaluation framework, while preserving the fixed reference defined in this paper. Each benchmark comes with a visualization interface, detailed descriptions of inputs and outputs, and instructions for accessing the data and constructing prediction files. Anyone can register and submit predictions, which are automatically ranked based on \texttt{total\_error} as defined in Section~\ref{sec:eval_metric}. See Section~\ref{app:bench} for additional details. The platform is already active, with over 80 community submissions at the time of writing, indicating early adoption by the community.

Several trends emerge from these results. MMGP is highly competitive when a meaningful alignment or fixed correspondence between samples is available, as in \texttt{Tensile2d} and \texttt{Rotor37}, but it is not directly applicable to datasets with variable topology such as \texttt{2D\_MultiScHypEl} and \texttt{2D\_ElPlDynamics}. MARIO performs particularly well on \texttt{Rotor37}, \texttt{2D\_profile}, and \texttt{VKI-LS59}, suggesting that implicit coordinate-based representations with geometric conditioning are well suited to smooth or shock-dominated CFD fields on varying geometries. Its weaker performance on \texttt{2D\_MultiScHypEl} indicates that topology changes and localized stress concentrations remain challenging for such latent geometric encodings. Vi-Transformer and Augur provide robust intermediate performance across all datasets, reflecting the flexibility of mesh partitioning and tokenization. FNO performs well on some lower-dimensional settings but degrades strongly on \texttt{Rotor37} and \texttt{2D\_profile}, where projection from anisotropic or three-dimensional mesh-based data to regular grids introduces both approximation error and computational overhead.

Overall, these results suggest that surrogate-model performance depends strongly on the match between data heterogeneity and architectural assumptions, and motivate benchmarks that preserve native mesh structure rather than reducing all datasets to regular grids.

\section{Conclusion and perspectives}
\label{sec:conclusion}
PLAID provides a unified data layer for heterogeneous physics simulation data, preserving the complexity of raw simulations while enabling interoperable datasets and reproducible evaluation under realistic industrial conditions. By unifying diverse datasets, storage backends, and benchmark protocols, PLAID turns heterogeneous simulations (with variable geometries, meshes, topologies, time dependence, and physical regimes) into reusable machine learning benchmarks. This work highlights that progress in scientific machine learning requires not only improved architectures, but also standardized data representations, reusable data infrastructure, and community-driven evaluation protocols.


\section*{Limitations}
The present release focuses on structural mechanics and CFD, with limited coverage of time-dependent problems. Broader physics domains, additional transient datasets, and time-resolved rollout diagnostics are left for future extensions. Test outputs are withheld to preserve leaderboard integrity, while training data, access code, scoring scripts, and benchmark entry points are released.

\section*{Impact Statement}
This work aims to advance machine learning for scientific and engineering simulation by improving dataset standardization, reproducible benchmarking, and interoperability across simulation and learning workflows. Its potential positive impacts include more transparent evaluation of surrogate models and broader reuse of physics simulation data. Possible risks include misuse of surrogate models outside their validated regimes, overreliance on benchmark scores, and unintended disclosure of sensitive industrial simulation data when releasing datasets. These risks are mitigated here by focusing on benchmark protocols, documented data representations, and controlled release of test outputs.

\section*{Software and Data}
All code, datasets, and benchmark entry scripts are provided (except for the commercial solution from Augur).
Datasets are distributed through Zenodo and Hugging Face, and the PLAID library provides the high-level accessors used throughout the paper. The benchmark code repository contains the scripts and configurations required to reproduce the reported results.

\bibliographystyle{icml2026}
\bibliography{references}

\clearpage
\appendix
\onecolumn

\section{Detailed benchmark results}
\label{app:benchdetails}

Detailed individual metrics for all methods and all datasets are provided in Table~\ref{tab:PLAID_benchmark}.

\begin{table}[!htbp]
\small
\centering
  \caption{RRMSE and total\_error on PLAID benchmarks, best on each line is {\bf bold}, second best is \underline{underlined}. Missing entries for MMGP correspond to non-applicable settings.}
 \begin{tabularx}{\textwidth}{L{3.12cm}C{1.8cm}C{1.8cm}C{1.8cm}C{1.8cm}C{1.8cm}C{1.8cm}}
  \toprule
 {\bf{Field}}, \textit{scalar} output & MGN  & MMGP & Vi-Transf. & Augur & FNO & MARIO \\
  \midrule
  {\texttt{Tensile2d}}  &  &  &  &  &  &  \\
  \midrule
  {\bf{U1}}  & 0.0788 &  \bf{0.0015} & 0.0086 & 0.0093 & 0.0174 & \underline{0.0023} \\
  {\bf{U2}}  & 0.1237 &  \bf{0.0009} & 0.0091 & 0.0135 & 0.0110  & \underline{0.0030}  \\
  {\bf{sig11}}  & 0.1726 &  \bf{0.0031} & 0.0184 & 0.0187 & 0.0250  & \underline{0.0040}  \\
  {\bf{sig22}}  & 0.0560 &  \bf{0.0013} & 0.0102 & 0.0099 & 0.0057  & \underline{0.0017}  \\
  {\bf{sig12}}  & 0.0570 &  \bf{0.0021} & 0.0146 & 0.0121 & 0.0135 & \underline{0.0023}   \\
  \textit{max\_von\_mises}  & 0.0185  & \bf{0.0050} & 0.0090 & 0.0219 & \underline{0.0085} & 0.0088  \\
  \textit{max\_U2\_top} & 0.0292  & \bf{0.0053} & 0.0203 & 0.0344 & 0.0152  & \underline{0.0063}  \\
  \textit{max\_sig22\_top} & 0.0030  & \bf{0.0017} & 0.0021 & 0.0030 & \underline{0.0021} & 0.0023   \\
  \midrule
  \bf{total\_error}  & 0.0673 & \bf{0.0026} & 0.0116 & 0.0154 & 0.0123 & \underline{0.0038}  \\
  \midrule
 \vspace{0.2em}
  {\texttt{2D\_MultiScHypEl}} &  &  &  &  & & \\
  \midrule
  {\bf{u1}}  & 0.0400 & - & 0.0173 & \underline{0.0140} & \bf{0.0115} & 0.0336  \\
  {\bf{u2}}  & 0.0444 & - & 0.0172 & \underline{0.0164} & \bf{0.0117} & 0.0377  \\
  {\bf{P11}}  & 0.0383 & - & \underline{0.0337} & \bf{0.0185} & 0.0353 & 0.0536  \\
  {\bf{P12}}  & 0.0670 & - & 0.0581 & \bf{0.0316} & \underline{0.0513} & 0.1067  \\
  {\bf{P22}}  & 0.0383 & - & \underline{0.0343} & \bf{0.0189} & 0.0359 & 0.0539  \\
  {\bf{P21}}  & 0.0663 & - & 0.0571 & \bf{0.0311} & \underline{0.0510} & 0.1053  \\
  {\bf{psi}}  & 0.0443 & - & \underline{0.0312} & \bf{0.0239} & 0.0329  & 0.0456  \\
  \textit{effective\_energy} & \bf{0.0111} & - & \underline{0.0113} & 0.0312 & 0.0120 & 0.0220 \\
  \midrule
  \bf{total\_error}  & 0.0437 & - & 0.0325 & \bf{0.0232} & \underline{0.0302} & 0.0573   \\
  \midrule
 \vspace{0.2em}
  {\texttt{2D\_ElPlDynamics}}  &  &  &  &  &  &  \\
  \midrule
  {\bf{U\_x}}  & 0.0195 & - & 0.0186 & 0.0264 &  \bf{0.0031} & \underline{0.0059}  \\
  {\bf{U\_y}}  & 0.2208 & - & \bf{0.0269} & 0.0427 &  \underline{0.0399} & 0.0580 \\
  \midrule
  \bf{total\_error}  & 0.1202 & - & \underline{0.0227} & 0.0346 &  \bf{0.0215} & 0.0319 \\
  \midrule
 \vspace{0.2em}
 {\texttt{Rotor37}} &  &  &  &  &  & \\
  \midrule
  {\bf{Density}}  & 0.0114 & \underline{0.0031} & 0.0063 & 0.0055 & 0.0840 & \bf{0.0026}  \\
  {\bf{Pressure}}  & 0.0114 & \underline{0.0030} & 0.0062 & 0.0053 & 0.0836 & \bf{0.0026}  \\
  {\bf{Temperature}} & 0.0024 & \underline{0.0008} & 0.0019 & 0.0012 & 0.0086 & \bf{0.0005}  \\
  \textit{Massflow}  & 0.0061 & \bf{0.0005} & \underline{0.0010} & 0.0028 & 0.0046 & 0.0019  \\
  \textit{Compression\_ratio} & 0.0060 & \bf{0.0005} & \underline{0.0011} & 0.0028 & 0.0042 & 0.0016  \\
  \textit{Efficiency}  & 0.0071 & \bf{0.0005} & \underline{0.0007} & \underline{0.0019} & 0.0031 & 0.0016 \\
  \midrule
  \bf{total\_error}  & 0.0074 & \bf{0.0014} & 0.0029 & 0.0033 & 0.0313 & \underline{0.0017}  \\
  \midrule
 \vspace{0.2em}
  {\texttt{2D\_profile}}   &  &  &  &  &  & \\
  \midrule
  {\bf{Mach}} & 0.0604 & 0.0439 & \underline{0.0359} & 0.0469 & 0.0988  & \bf{0.0355}   \\
  {\bf{Pressure}} & 0.0466 & 0.0208 & \bf{0.0170} & 0.0248 & 0.0785 & \underline{0.0196}  \\
  {\bf{Velocity-x}} & 0.0735 & 0.0471 & \bf{0.0408} & 0.0538 & 0.1148 & \underline{0.0460}  \\
  {\bf{Velocity-y}} & 0.0566 & 0.0342 & \underline{0.0309} & 0.0445 & 0.0967 & \bf{0.0215}  \\
  \midrule
  \bf{total\_error}  & 0.0593 & 0.0365 & \underline{0.0312} & 0.0425 & 0.0972 & \bf{0.0307}  \\
  \midrule
 \vspace{0.2em}
  {\texttt{VKI-LS59}}   &  &  &  &  &  & \\
  \midrule
  {\bf{nut}}  & 0.1656 & 0.0822 & 0.0498 & \underline{0.0483} & 0.0846 & \bf{0.0259}  \\
  {\bf{mach}}  & 0.0451 & 0.0309 & 0.0232 & 0.1896 & \underline{0.0180} & \bf{0.0112}  \\
  \textit{Q}  & 0.0716 & \bf{0.0023} & 0.0052 & 0.0074 & \underline{0.0047} & 0.0052  \\
  \textit{power}  & 0.0403 & \bf{0.0057} & 0.0083 & 0.0070 & \underline{0.0062} & 0.0077  \\
  \textit{Pr}  & 0.0068 & 0.0026 & 0.0024 & 0.0040 & \underline{0.0019} & \bf{0.0018}  \\
  \textit{Tr}  & 0.0001 & \bf{0.0000} & \bf{0.0000} & \bf{0.0000} & \bf{0.0000} & \bf{0.0000}  \\
  \textit{eth\_is}  & 0.1912 & 0.1224 & 0.0621 & 0.1228 & \underline{0.0539} & \bf{0.0453}  \\
  \textit{angle\_out}  & 0.0263 & 0.0033 & 0.0031 & 0.0048 & \underline{0.0027} & \bf{0.0023}  \\
  \midrule
  \bf{total\_error}  & 0.0684 & 0.0312 & \underline{0.0193} & 0.0267 & 0.0215 & \bf{0.0124} \\
  \midrule
 \end{tabularx}
 \label{tab:PLAID_benchmark}
\end{table}

\section{Details on the ML models used in the benchmark}
\label{app:mldetails}

We briefly present the main competing methods used in the benchmark, together with practical details about their implementation. Readers are encouraged to refer to the original publications for a complete description of each method.

All methods are trained on the same dataset splits and evaluated using the standardized PLAID benchmark protocol described in Section~\ref{sec:benchmark}. Full implementation details, including complete hyperparameter configurations, training scripts, and data preprocessing pipelines, are provided in the accompanying  \href{https://github.com/PLAID-lib/plaid-benchmarks}{benchmark-code repository}, ensuring full reproducibility of all experiments (except for the commercial solution from Augur).

We briefly highlight how each method handles data heterogeneity, which is a central challenge addressed in this work.

\subsection{MGN}

MeshGraphNet (MGN) operates directly on mesh-based representations, enabling it to naturally handle geometric variability and unstructured discretizations.

\subsubsection{Method}

\citet{pfaff2021learning} introduced MGN, a framework designed for learning mesh-based simulations using graph neural networks. The model is capable of being trained to simulate dynamic solutions by passing messages over a meshed domain, predicting acceleration at each mesh node at a given time step. This prediction allows for the calculation of the output field at the next time step through forward integration. Specifically, MGN is trained using one-step supervision and can be applied iteratively to generate long trajectories during inference. The architecture of MeshGraphNet is composed of encoding, processing, and decoding steps. In this work, MGN has been adapted to predict steady-state fields.

\begin{figure}[!htbp]
\centering
\includegraphics[width=0.8\textwidth]{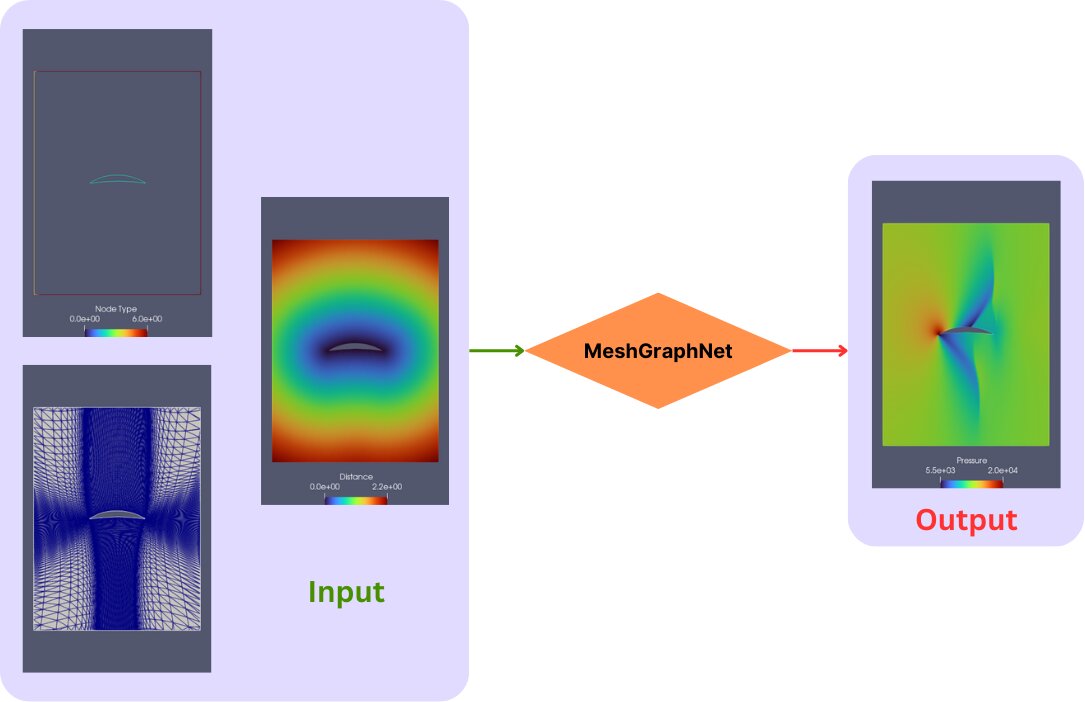}
\caption{Illustration of MGN workflow to predict steady-state pressure field of a sample from the \texttt{2D\_profile} dataset.}
\label{fig:workflow_mgn}
\end{figure}

We utilize the following features as input (see Figure \ref{fig:workflow_mgn} for the workflow diagram):

\begin{itemize}
    \item the distance of each node to the boundary,
    \item the type of node,
    \item the coordinates of the node.
\end{itemize}

\subsubsection{Experiments}

In this section, we provide a summary of the experiments conducted on various datasets.

For all datasets, we trained two separate models: one focused on field predictions and the other on scalar predictions. For scalar outputs, a readout layer inspired by~\citet{kipf2017semisupervisedclassificationgraphconvolutional} is added to the model. Except for the \texttt{2D\_profile} and \texttt{2D\_ElPlDynamics} datasets, we only required a single model since it does not include scalar prediction tasks.

The LeakyReLU is chosen as the activation function, and all models are trained for 1000 epochs, except the \texttt{2D\_ElPlDynamics} dataset with 100 epochs.

The input node features consist of those introduced in the previous section, combined with input scalars if they exist. Given two node coordinates $x_i$ and $x_j$, the calculation for edge features is based on $\exp(-||x_i - x_j||^2_2/(2h^2))$, where $h$ represents the median value of the edge lengths within the mesh.

The rest of architecture details and training information are outlined in Table~\ref{tab:mgn_fields} and Table \ref{tab:mgn_scalars}.

\begin{table}[!htbp]
\small
\centering
\caption{Field MGN: Architecture details and training statistics across datasets.}
\begin{tabularx}{\textwidth}{L{2.5cm} C{2.25cm} C{1.87cm} C{1.87cm}  C{1.87cm} C{1.87cm}  C{1.87cm} }
  \toprule
  \texttt{Dataset} & Message Passing Steps & Latent Size & Nbe epochs & Batch size & Training Time & Hardware \\
  \midrule
  \texttt{Tensile2d}  & 10 & 16 &  1000  & 1 &  3h46min  & 1 $\times$ A100 \\
  \texttt{2D\_MultiScHypEl}  & 10 & 32 &  1000  & 1 &  5h54min  & 1 $\times$ A100  \\
  \texttt{Rotor37}  & 10 & 64 &  1000  & 1 &  19h24min  & 1 $\times$ A100  \\
  \texttt{2D\_profile}  & 10 & 128 &  1000 & 1 &  17h27min  & 1 $\times$ A100 \\
  \texttt{VKI-LS59}  & 10 & 64 &  1000 & 1 & 16h32min & 1 $\times$ A100 \\
  \texttt{2D\_ElPlDynamics}  & 15 & 64 & 100 & 8 & 15h43min & 8 $\times$ A100 \\
  \bottomrule
\end{tabularx}
\label{tab:mgn_fields}
\end{table}

\begin{table}[!htbp]
\small
\centering
\caption{Scalar MGN: Architecture details and training statistics across datasets. The \texttt{2D\_profile} and \texttt{2D\_ElPlDynamics} benchmarks do not include scalar outputs.}
\begin{tabularx}{\textwidth}{L{2.5cm} C{2.25cm} C{1.87cm} C{1.87cm}  C{1.87cm} C{1.87cm}  C{1.87cm} }
  \toprule
  \texttt{Dataset} & Message Passing Steps & Latent Size & Nbe epochs & Batch size & Training Time & Hardware \\
  \midrule
  \texttt{Tensile2d}  & 10 & 32 &  1000  & 1  &  4h6min  & 1 $\times$ A100 \\
  \texttt{2D\_MultiScHypEl}  & 10 & 16 &  1000  & 1 &  6h  & 1 $\times$ A100  \\
  \texttt{Rotor37}  & 10 & 16 &  1000  & 1 &  10h  & 1 $\times$ A100  \\
  \texttt{VKI-LS59}  & 10 & 16 &  1000 & 1 & 9h13min & 1 $\times$ A100 \\
  \bottomrule
\end{tabularx}
\label{tab:mgn_scalars}
\end{table}

\subsection{MMGP}

The Mesh Morphing Gaussian Process (MMGP) method relies on mesh morphing to align geometries, which limits its applicability in the presence of topological variations.

\subsubsection{Method}

We refer the reader to~\cite{casenave2024mmgp} for a complete presentation of the method. MMGP combines four main ingredients: (i) mesh morphing, (ii) finite element interpolation, (iii) dimensionality reduction, and (iv) Gaussian process regression. Together, these enable learning mappings between geometries and solution fields for PDEs, even when the input geometry is provided as non-parametrized meshes.

\begin{figure}[!htbp]
\centering
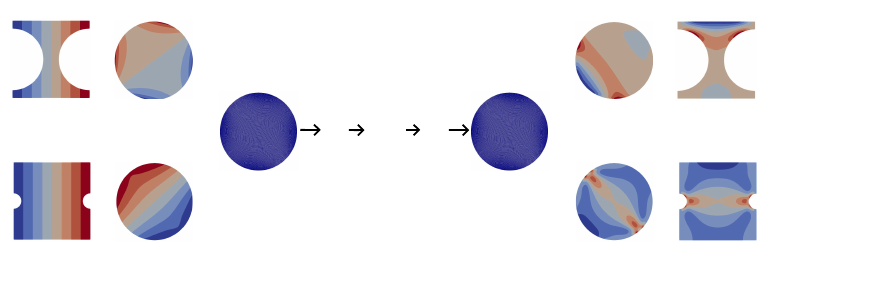
\caption{Illustration of the MMGP inference workflow for the prediction of an output field of interest~\cite{casenave2024mmgp}.}
\label{fig:mmgp}
\end{figure}

An overview of the workflow is illustrated in Figure~\ref{fig:mmgp}, which should be read from left to right. On the left are sample-specific input geometries; on the right are the corresponding solution fields defined on these geometries.

Since input meshes are not parametrized, they must first be embedded into a learnable space. MMGP does this by interpreting mesh vertex coordinates as continuous fields (e.g., the $x$-coordinate field shown in the left column of Figure~\ref{fig:mmgp}, exhibiting vertical iso-lines). Each mesh is then deterministically morphed onto a reference geometry—the unit disk in this 2D example, but it can be one of the training samples shape. Next, each sample morphed coordinate fields are projected onto a common mesh of the reference geometry via finite element interpolation. This ensures all samples share a consistent discretization, making them compatible with standard dimensionality reduction techniques like PCA. The result is a compact, fixed-size representation of the geometry. When scalar inputs are present, they can be concatenated to the reduced vector.

A similar procedure is applied to the output fields: morphing onto the reference geometry, projection onto the common mesh, and PCA compression yield low-dimensional field representations aligned with the geometric embeddings.

These deterministic preprocessing steps transform the original complex problem—mapping between high-dimensional and irregularly discretized fields—into a standard regression task between low-dimensional vectors. This enables the use of classical regression models; we adopt Gaussian process regression due to its robustness, accuracy, and built-in uncertainty quantification.

MMGP offers several practical advantages: it handles large meshes efficiently, produces interpretable models, and delivers high accuracy in our experiments, with uncertainty estimates. In industrial design applications, where data can lie on low-dimensional manifolds, small models like MMGP can be especially effective—provided that the reparametrization (or embedding) is constructed appropriately, here with the morphing.

The main limitations of MMGP are tied to the morphing step, which currently requires problem-specific setup, and the fact that morphing and interpolation must still be performed at inference time. These challenges are addressed in recent works~\cite{KABALAN2025115929, kabalan2024morphing}, which introduce automatic alignment and online-efficient morphing strategies. \citet{kabalan2025ommgp} proposed improvements, where optimization techniques are used to generate morphings that maximize PCA compression.

All mesh and field operations are implemented using the Muscat library~\cite{muscat2023, Bordeu2023}. An upcoming release will include a GPU-accelerated finite element interpolation routine, significantly improving inference latency.

Additional improvements of MMGP are possible, by replacing the linear decoder of the PCA by a non-linear one that accounts for high-order interactions among the selected POD modes and includes a rotation of the POD basis and a polynomial correction, as proposed by \citet{geelen2023learning}.

Physics-based models compatible with the morphing, finite element interpolation and dimensionality reduction of MMGP have been proposed. The physics equation can be efficiently assembled and solved on the low-dimension space spanned by the PCA modes obtained after morphing, instead of using data-driven low-dimensional models. In~\cite{BARRAL2024112727}, a hyper-reduced least-square Petrov-Galerkin scheme is used to reduce the Navier-Stokes equations, with morphing. While much more complicated to utilize, we expect such methods to greatly improve the accuracy, with a moderate additional computation cost.

\subsubsection{Experiments}

Hyperparameters and training statistics for the MMGP experiments are listed in Table~\ref{tab:MMGP_res}. We first mention that MMGP has not been applied to the \texttt{2D\_ElPlDynamics} and \texttt{2D\_MultiScHypEl} datasets, since the method is yet to be extended to variable topology settings.

We notice that \texttt{Rotor37} and \texttt{VKI-LS59} do not require morphing, since the samples' meshes have the same number of nodes. In \texttt{Tensile2d} and \texttt{2D\_profile}, systematic morphing strategies to align the shapes are sufficient, with respectively Tutte barycentric embedding~\cite{casenave2024mmgp} (Annex B) and elasticity-based automatic morphing~\cite{KABALAN2025115929}.

Since the \texttt{VKI-LS59} dataset exhibits discontinuities due to the presence of shock waves, a non-linear decoder \cite{geelen2023learning} was employed to reconstruct the fields of interest.
For the compression of the \texttt{mach} fields, 5 POD modes and a polynomial order of 3 were used, while 40 POD modes were retained for the compression of the \texttt{nut} fields. Since polynomial decoders are prone to overfitting, the number of modes and the polynomial order were selected through a $k$-fold cross-validation procedure on the training set.

Since the solution fields of \texttt{2D\_profile} and \texttt{VKI-LS59} feature complex structures (e.g. shocks of variable position), we expect the involved optimal morphing strategy of~\citet{kabalan2025ommgp} to significantly improve the results of MMGP on these cases.


\begin{table}[!htbp]
\small
\centering
  \caption{Hyperparameters and training statistics for the MMGP experiments (on an AMD EPYC 9534 CPU). Training times include all preprocessing (morphing, finite element interpolation and dimensionality reduction), in addition to the training of the Gaussian processes. *For \texttt{VKI-LS59}, X-Y stands for the number of modes and polynomial order of the decoder for the \texttt{mach} and \texttt{nut} fields respectively. **Not including morphing time (which takes approximately 10min on 300 cores).}
 \begin{tabularx}{\textwidth}{L{1.87cm} L{2.75cm} C{1.87cm} C{1.87cm} C{1.87cm} C{1.87cm} C{1.87cm}}
  \toprule
 \texttt{Dataset} & Morphing & PCA modes (shape) & PCA modes (field) & GP kernel & Training time & Hardware\\
  \midrule
  \small{\texttt{Tensile2d}}  & Tutte & 8 & 8 & Mat\'{e}rn 5/2 & 13min02s & 128 cores \\
  \small{\texttt{Rotor37}}  & None & 32 & 64 & Mat\'{e}rn 5/2 & 6min13s & 128 cores \\
  \small{\texttt{2D\_profile}} & Elasticity & 16 & 32 & RBF & 18min32s** & 12 cores \\
  \small{\texttt{VKI-LS59}} & None & 13 & 5-3/40-1* & Mat\'{e}rn 5/2 & 4min13s & 64 cores \\
  \bottomrule
 \end{tabularx}
 \label{tab:MMGP_res}
\end{table}

\subsection{Vi-Transformer and Augur}

These methods rely on mesh partitioning and tokenization, enabling flexible handling of irregular geometries while preserving local structural information.

\subsubsection{Method}

\paragraph{Transformers for long context range regression.}
The natural way of dealing with mesh-based regression problems is to use GNN models which rely on message-passing. Although these are great at capturing information locally, they struggle to retrieve it at long distances. Indeed, the smallest number of GNN layers needed to have a receptive field that covers the whole graph is half the diameter of the graph. This becomes computationally impractical in the context of large simulation meshes.
This behavior is analogous to Convolutional Neural Networks (CNNs) in Computer Vision (CV) where long-range dependencies are only captured at the deeper levels of the network.
One way of alleviating this is to consider transformer architectures, which compute similarities between all the input tokens simultaneously thanks to the attention mechanism, thus removing the need to have infeasibly deep networks.

\paragraph{Transformers on very large data.}
One of the main challenges of transformers in this case is to handle the large size of the meshes (in the order of tens of thousands of points per mesh, and up to millions with practical industrial problems).
Currently, the computational bottleneck of transformers is a widely considered subject: given $N$ tokens of dimension $D$, the critical issue of self attention is that one needs $N^2 \times \mathcal{D}$ operations where $\mathcal{D}\approx D$ is the size of the embedding of each token, and $N^2$ is the cost to compute the Gram matrix of the $N$ tokens (this computation cost is also a memory one as storing the matrix requires also $N^2\mathcal{D}$ numbers).

Many papers have focused on the possibility to linearize the cost of self-attention, for example:
\begin{itemize}
    \item \citet{kitaev2020reformer} introduces Reformer which considers the formulation of the attention mechanism : $\mathrm{softmax}\left(\frac{QK}{\sqrt{D}}\right)$ with the key and query matrices (respectively $K$ and $Q$), capitalizing on the fact that for a given query $Q_i$, only the keys which provide high dot products with $Q_i$ will have a significant impact on the value of $\mathrm{softmax}\left(\frac{Q_iK^T}{\sqrt{D}}\right)$.
    Therefore, Reformer makes use of locality-sensitive hashing for only computing the $Q_iK^T_j$ products with the $p$ keys that are closest to a query, where $p \in \mathbb{N}$ is a chosen hyperparameter, efficiently linearizing the self attention.
    \item \citet{wang2020linformer} introduces Linformer. Coarsely, Linformer relies on the Nyström approximation to approximate the Gram matrix of self attention. Precisely, while the Nyström approximation replaces an $n\times n$ symmetric matrix $A$ by $UU^T$ where $U$ is only $n\times k$  containing the eigenvectors of largest eigenvalue, Linformer offers to learn $E,F$ such that $\mathrm{softmax}\left(\frac{QK}{\sqrt{D}}\right)\approx EF^T$.
    This also offers a linear approximation of the self attention computation.
\end{itemize}

This has also been tackled in CV tasks \cite{dosovitskiy2021imageworth16x16words}, where self-attention is not applied on pixels directly but on pixel-patches that aggregate pixel neighborhoods into tokens, thus drastically reducing the self-attention's input sequence length.

\paragraph{Transformers for large scale point-wise regression.}
The most used transformer architectures are in one of two categories. The auto-regressive sequence-to-sequence transformers, mostly used in Natural Language Processing (NLP) for text generation, and the sequence-to-class ones which are used both in NLP, as in sentiment analysis \cite{Durairaj2021}, spam detection \cite{jamal2023improvedtransformerbasedmodeldetecting}, long document classification \cite{dai2022revisitingtransformerbasedmodelslong}, and CV with image classification \cite{dosovitskiy2021imageworth16x16words}.

Both are quite different from the point-wise regression objective of the PLAID benchmarks. Indeed, the first method generates new token sequences of arbitrary lengths, while the second only makes use of transformer encoders with neural network heads to obtain a probability distribution on a set of classes.

Some work has been conducted in order to tackle regression problems with transformers:
\begin{itemize}
    \item Segformer \cite{xie2021segformersimpleefficientdesign} addresses this in the case of image segmentation; it uses a multiscale U-type transformer to sequentially downscale the input image, and uses a multiscale MLP head to decode these downscaled states into the output segmentation mask.
    \item Point Transformer \cite{zhao2021pointtransformer} also uses a U-style encoder-decoder architecture, this time on 3D point-cloud data for both segmentation and classification.
    \item TransCFD \cite{JIANG2023106340} tackles airfoil surrogate CFD modeling by using a decoder-only architecture from a latent embedding of the input geometry. It relies on structured regular grids (images) of the inputs, and not arbitrary mesh discretizations.
    \item Point Transformer V3 \cite{wu2024pointtransformerv3simpler} groups points together and computes attention scores within these groups. Local and long-distance information are captured through different serializations of the input mesh.
\end{itemize}

Both Segformer and TransCFD make use of the regular nature of their data to precisely decode (and/or encode) the output (and/or input) fields. Point Transformers, on the other hand, handle unstructured point-cloud data. Although these methods fit the nature of the PLAID benchmark, we propose lighter methods that stick more closely to the classical transformer model.

\begin{figure}[!htbp]
    \centering
    \includegraphics[width=0.8\linewidth]{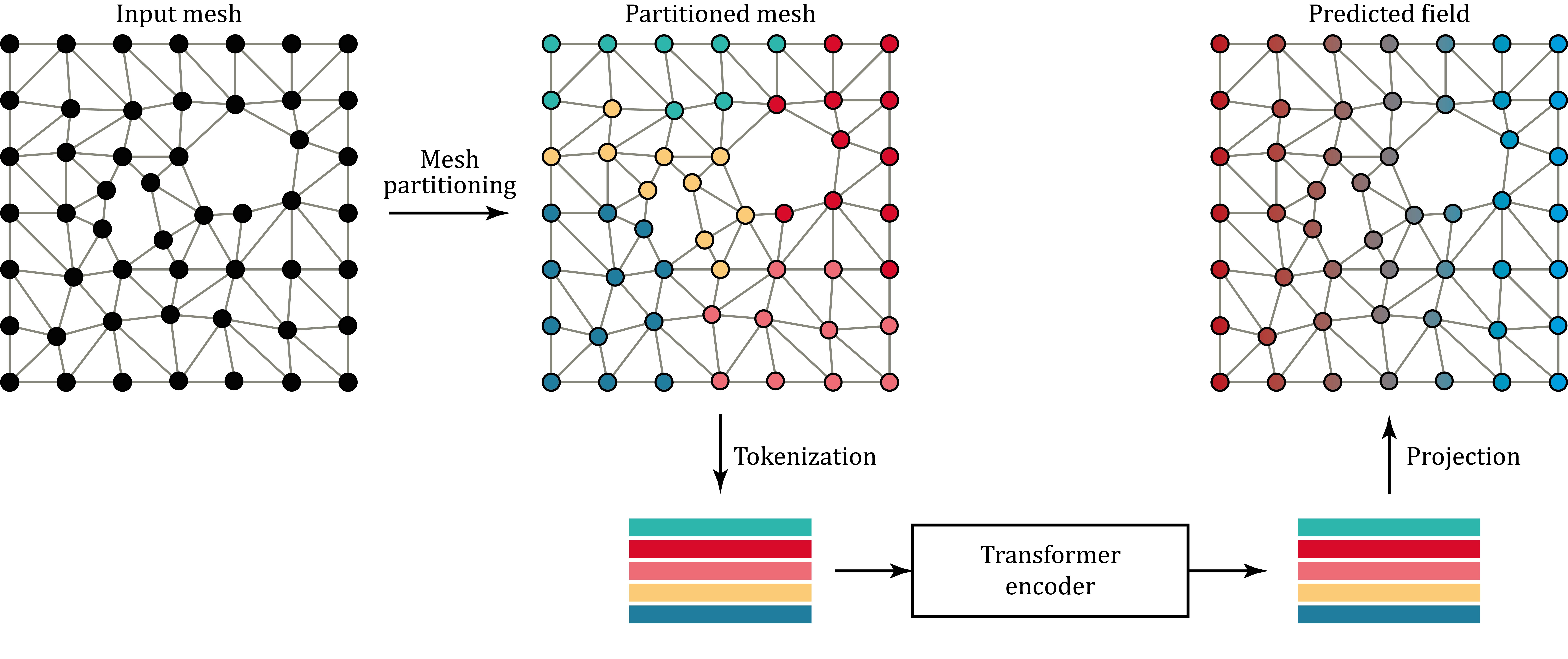}
    \caption{Vi-Transformer architecture. Input meshes are partitioned using the Metis domain decomposition algorithm \cite{metis}. Each such sub-domain is then tokenized before passing through the transformer encoder. In the end, each token is decoded into its domain's corresponding fields. Input scalars are embedded during the tokenization procedure while output scalars are estimated as uniform fields.}
    \label{fig:ViT}
\end{figure}

\paragraph{Vi-Transformer for mesh field regression.}
The chosen approach relies on a transformer encoder architecture and is analogous to Vision Transformers (Vi-Transformer). Rather than considering each node of the mesh as a token by its own, the encoder takes as input tokenized point-cloud patches. Local information is kept within the patches while long-range information is retrieved through the transformer's mapping, which compares all token pairs together. The general architecture of the Vi-Transformer is depicted in Figure \ref{fig:ViT}. Although this model primarily focuses on steady-state simulations, it can be trained in an auto-regressive way to learn unsteady simulation increments; and reconstructing, on inference, the full evolution by auto-regressively estimating the considered fields.

\paragraph{Augur Transformer model.}
Augur has developed Transformer models specifically designed for numerical simulations. These models share fundamental architectural similarities with Vision Transformers (ViT), where the computational mesh is decomposed into patches. Each patch is embedded into a latent space, resulting in the input tokens for the Transformer architecture. This approach enables information exchange between local patches across long spatial distances, similar to how ViTs process image data.

\begin{figure}
    \centering
    \includegraphics[width=0.9\linewidth]{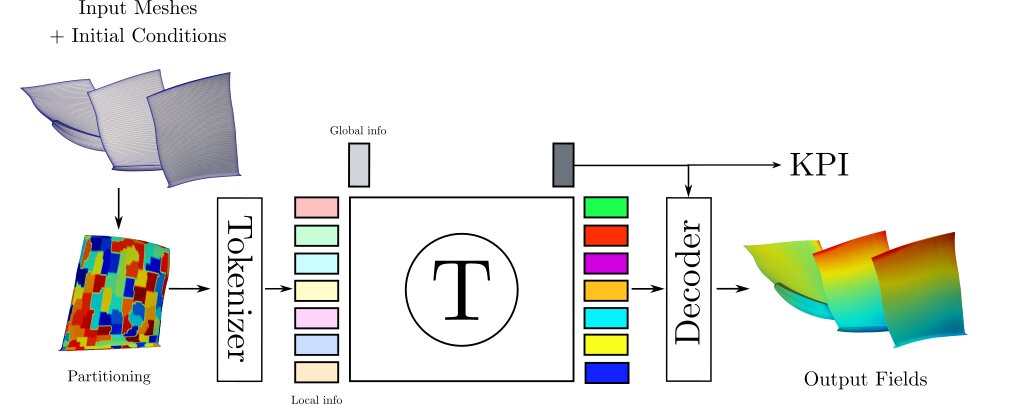}
    \caption{Augur Transformer architecture: Input meshes are partitioned using the Metis domain decomposition algorithm. Each subdomain is then tokenized before being passed through the Transformer. An additional global tensor is added to the Transformer to gather global information. Output fields are reconstructed using a decoder that leverages both local and global information. Output scalars (KPIs) are predicted directly from the global tensor.}
    \label{fig:augur_figure}
\end{figure}

The key innovation in Augur's approach lies in the decoding mechanism, addressing a critical question: how to properly reconstruct the output field from the processed sequence of tokens? In traditional ViT architectures, direct reconstruction from individually processed tokens can result in discontinuities at patch interfaces due to insufficient global context integration. Augur models overcome this limitation by incorporating a global information vector that aggregates data from all tokens. The decoder then uses a combination of point-specific information, processed local features, and global context to produce a more robust and consistent output field. Furthermore, unlike ViTs, Augur models do not treat scalar predictions as constant fields but instead derive them directly from the global information vector, enhancing prediction accuracy. The general architecture of the Augur model is depicted in Figure \ref{fig:augur_figure}.

\subsubsection{Experiments}

Both the Vi-Transformer and Augur models rely on a relatively small number of hyperparameters. These include the patch size (i.e., the number of nodes per patch), the latent dimension onto which the aggregated patches are projected, and the Transformer encoder hyperparameters, such as the number of heads, the number of transformer encoder layers and the dimension of the feedforward layer. Table \ref{tab:ViT_res} details the hyperparameters for the Vi-Transformer, while Table \ref{tab:augur_res} outlines those for the Augur model.

\begin{table}[!htbp]
\small
\centering
  \caption{(Vi-Transformer) Hyperparameters and training statistics for the Vi-Transformer experiments. Training times include all preprocessing (domain decomposition, tokenization), in addition to the training of the model itself. The number of attention heads is kept at 16 for all experiments.}
  \begin{tabularx}{\textwidth}{L{2.75cm} C{1.87cm} C{1.87cm} C{1.87cm} C{1.87cm} C{1.87cm} C{1.87cm}}
  \toprule
 \texttt{Dataset}                       & Patch size    & Latent dimension    & Feedforward dimension   & Nb encoder layers     & Training time     & Hardware\\
  \midrule
  \small{\texttt{Tensile2d}}            & 10            & 1024          & 512               & 5         & 2h43      & 1 $\times$ A30 \\
  \small{\texttt{2D\_MultiScHypEl}}     & 10            & 1024          & 512               & 5         & 2h07      & 1 $\times$ A30 \\
  \small{\texttt{Rotor37}}              & 30            & 1024          & 512               & 5         & 3h59      & 1 $\times$ A30 \\
  \small{\texttt{2D\_profile}}          & 25            & 1024          & 512               & 5         & 6h31      & 1 $\times$ A30 \\
  \small{\texttt{VKI-LS59}}             & 20            & 1024          & 512               & 5         & 6h39      & 1 $\times$ A30 \\
  \small{\texttt{2D\_ElPlDynamics}}     & 100           & 256           & 128               & 2         & 48min         & 1 $\times$ A30 \\
  \bottomrule
 \end{tabularx}
 \label{tab:ViT_res}
\end{table}

\begin{table}[!htbp]
\small
\centering
  \caption{(Augur) Hyperparameters and training statistics for the Augur experiments. Training times include all preprocessing (domain decomposition, tokenization), in addition to the training of the model itself.}
 \begin{tabularx}{\textwidth}{L{2.75cm} C{1.62cm} C{1.75cm} C{1.75cm} C{1.87cm} C{1.5cm} C{2.75cm}}
  \toprule
 \texttt{Dataset}                       & Patch size    & Latent dimension    & Feedforward dimension   & Nb encoder layers   & Training time     & Hardware\\
  \midrule
  \small{\texttt{Tensile2d}}            & 16            & 512          & 2048              & 8        & 1h11              & 1 $\times$ RTX 2080Ti \\
  \small{\texttt{2D\_MultiScHypEl}}     & 4            & 128           & 512               & 8        & 7h48              & 1 $\times$ RTX 2080Ti \\
  \small{\texttt{Rotor37}}              & 32           & 256           & 1024               & 8        & 2h30            & 1 $\times$ RTX 2080Ti \\
  \small{\texttt{2D\_profile}}       & 128           & 256           & 512               & 4        & 1h51            & 1 $\times$ RTX 2080Ti \\
  \small{\texttt{VKI-LS59}}              & 64           & 512           & 2048               & 4        & 2h15            & 1 $\times$ RTX 2080Ti \\
  \small{\texttt{2D\_ElPlDynamics} }              & 256           & 256          & 4               & 4        & 1h30            & 1 $\times$ RTX 2080Ti \\
  \bottomrule
 \end{tabularx}
 \label{tab:augur_res}
\end{table}

\subsection{FNO}

Fourier Neural Operators (FNO) operates on regular grids and therefore requires projection from unstructured meshes, which may introduce approximation errors in highly heterogeneous settings.

FNO belong to the class of operator learning architectures, i.e., they aim at approximating mappings between function spaces. By leveraging the Fast Fourier Transform (FFT) within their layers, FNOs represent functions on regular grids, effectively working in a finite-dimensional discretized space. This architecture enables learning transformations in the frequency domain, which has been shown to be particularly effective for modeling complex physical systems compared to standard convolutional neural networks.

\subsubsection{Method}

We use the classical FNO architecture introduced by~\citet{li2020fourier}. Let $u: \R^2 \rightarrow \R^k$ denote the input field. Each FNO layer is defined as the sum of a local linear transformation and a non-local integral operator parameterized in the Fourier domain:
\begin{align}
\mathcal{J}[u](x) = \sigma \left( W u(x) + \mathcal{F}^{-1} \left( W^* \mathcal{F}[u] \right)(x) \right),
\end{align}
where $W \in \R^{k \times k}$ and $W^*$ are learnable parameters, $\mathcal{F}$ and $\mathcal{F}^{-1}$ denote the Fourier transform and its inverse, and $\sigma$ is a pointwise non-linear activation function. Stacking multiple such layers yields a global operator that captures long-range dependencies through spectral convolution.

A key limitation of FNO models is that they operate on regular grids due to the use of FFT. As a result, applying FNOs to data defined on unstructured meshes requires a preprocessing step that projects the fields onto a regular grid, followed by a postprocessing step mapping predictions back to the original mesh. In our experiments, these projection operations are performed using Muscat~\cite{muscat2023,Bordeu2023}.

\subsubsection{Experiments}

\subsubsection{\texttt{2D\_ElPlDynamics}}

FNO can be used to model transient dynamics in physics-based systems. Among the datasets introduced in Section~\ref{sec:datasets}, \texttt{2D\_ElPlDynamics} is the only one featuring time-dependent behavior with evolving topology.

\paragraph{Training procedure.}
The training is performed in an autoregressive manner: given the input fields at time $t$, the model predicts the fields at time $t + dt$, similarly to an explicit time integration scheme. Once trained, full trajectories are obtained by recursively applying the model starting from the initial condition.

The input features consist of the displacement fields provided by the dataset (\texttt{U\_x}, \texttt{U\_y}), augmented with spatial coordinate fields ($x$, $y$), which are commonly used in FNO-based models to encode positional information.

As FNO operates on regular grids, the fields defined on unstructured meshes are first projected onto a regular grid before being processed by the network. After prediction, the output fields are projected back onto the original mesh to enable comparison with reference solutions. We summarize the input/output quantities in Table~\ref{tab:learning_features}.

\newlength{\widthfigures}
\setlength{\widthfigures}{0.265\textwidth}

\begin{table}[!htbp]
\centering
  \caption{Features throughout the learning process. The simulation at $t$ (column 1) is projected to a regular grid (column 2), which serves as input to the FNO model. The model predicts the fields at $t+dt$ (column 3), which are then mapped back to the original mesh.}
\begin{tblr}{
colspec = {Q[c]Q[c]Q[c]Q[c]},
rowspec={|Q[m]|Q[m]Q[m]Q[m]Q[m]|},
stretch = 0,
rowsep = 3pt,
}
      Attribute & Simulation at $t$ & Input FNO & Output FNO \\
      Mesh & \includegraphics[width=\widthfigures, valign=c]{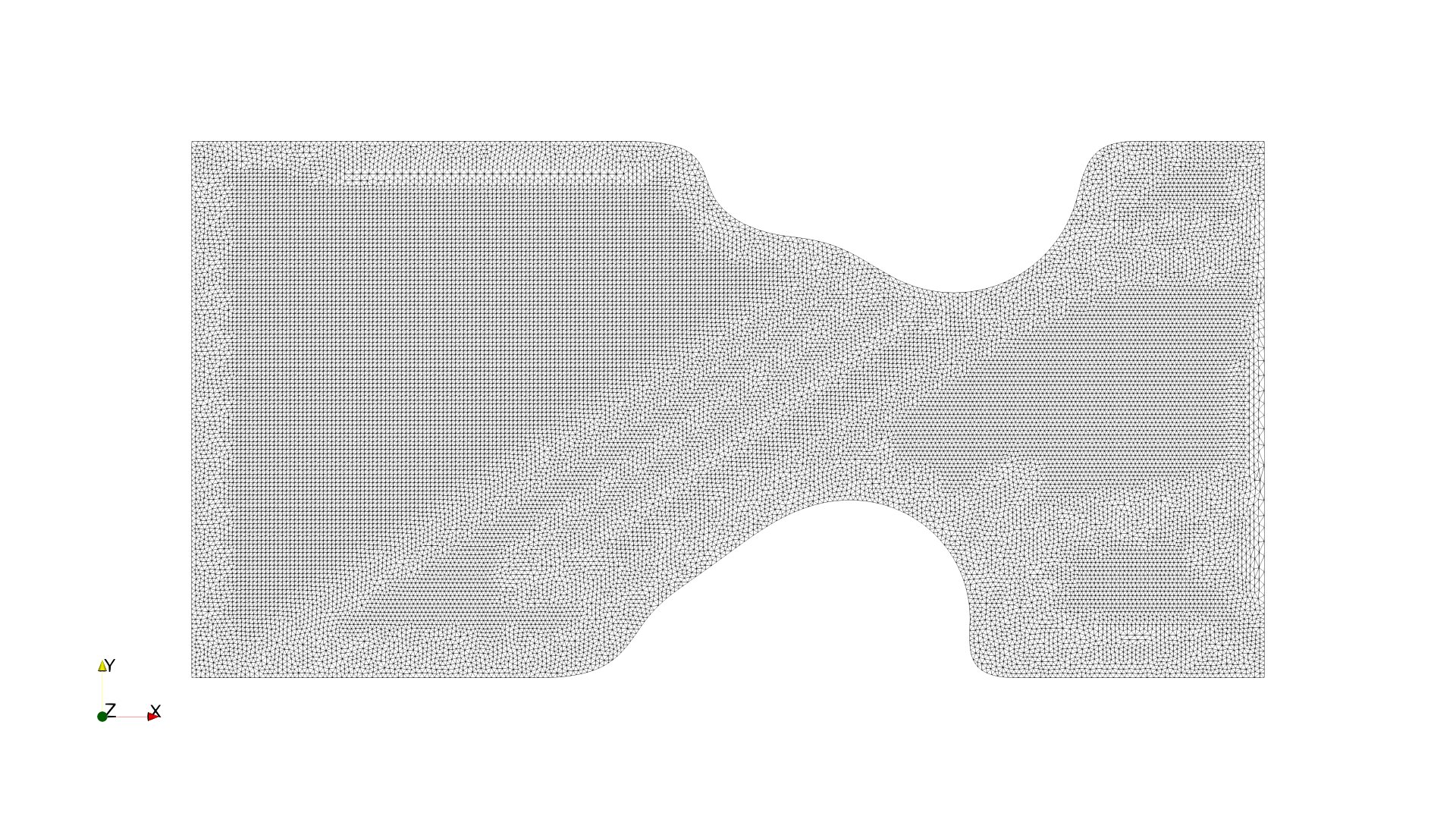} & - & - \\
      \texttt{U\_x} & \includegraphics[width=\widthfigures, valign=c]{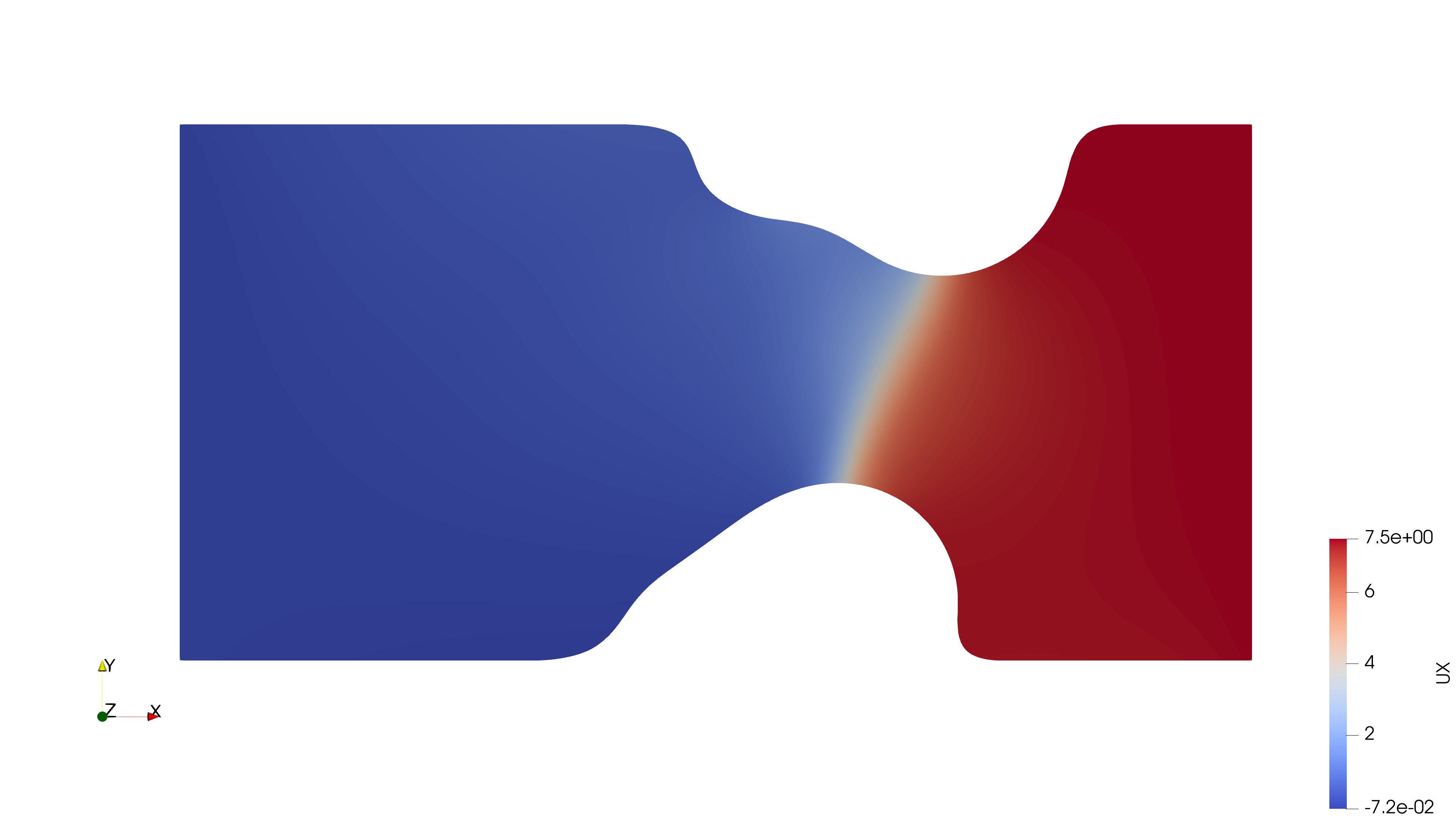} &  \includegraphics[width=\widthfigures, valign=c]{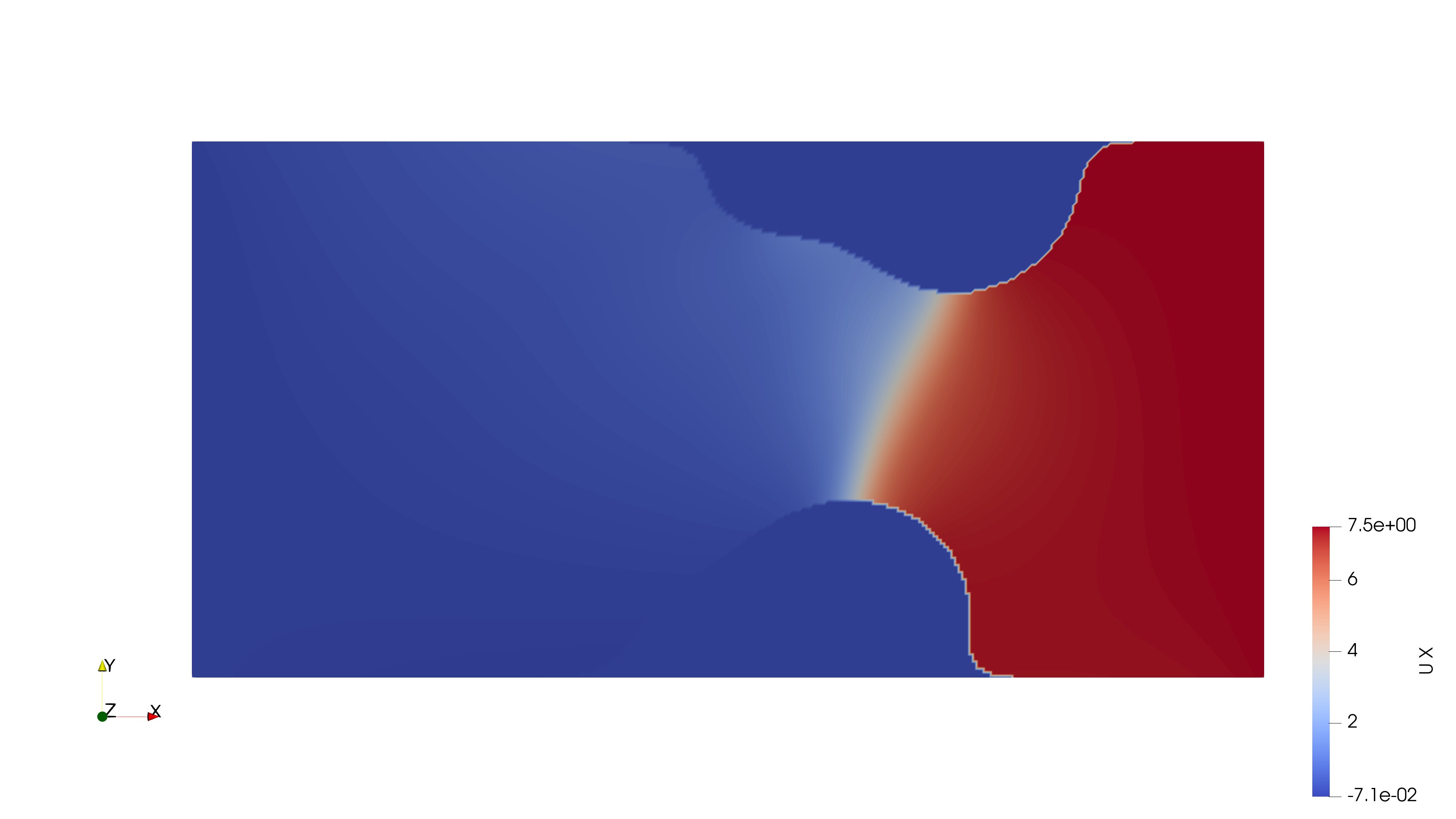} & \includegraphics[width=\widthfigures, valign=c]{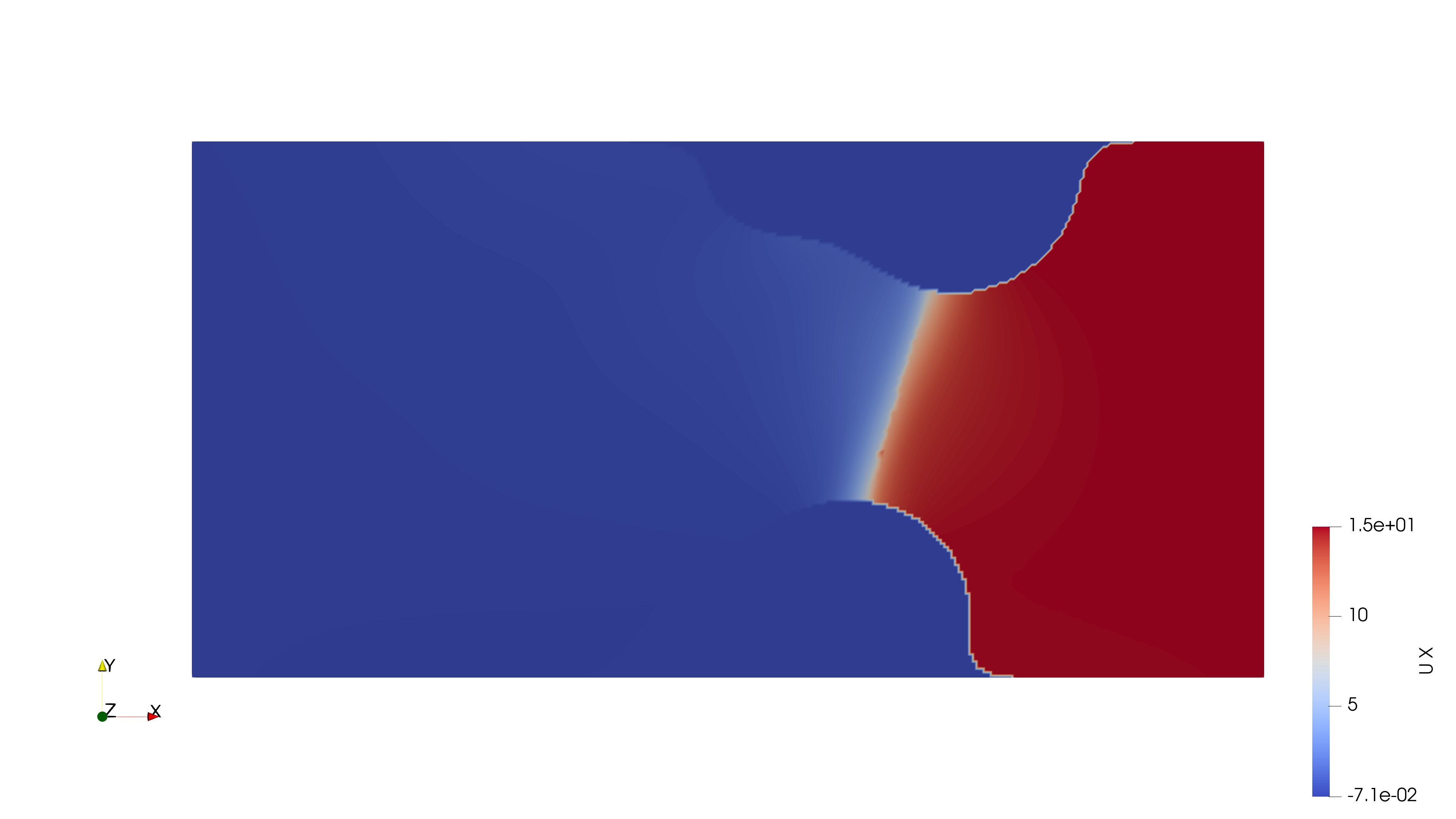}\\
      \texttt{U\_y}& \includegraphics[width=\widthfigures, valign=c]{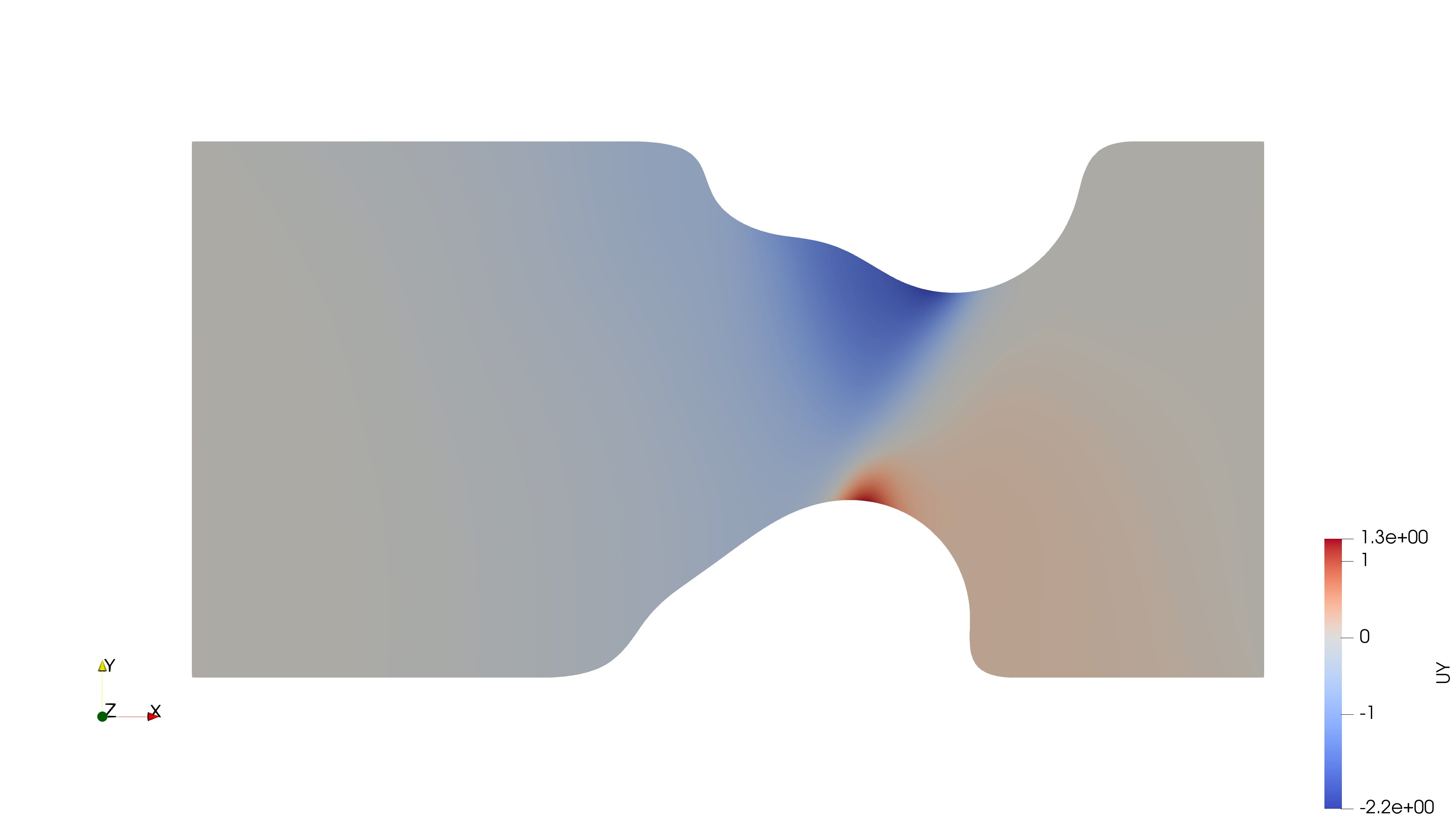} & \includegraphics[width=\widthfigures, valign=c]{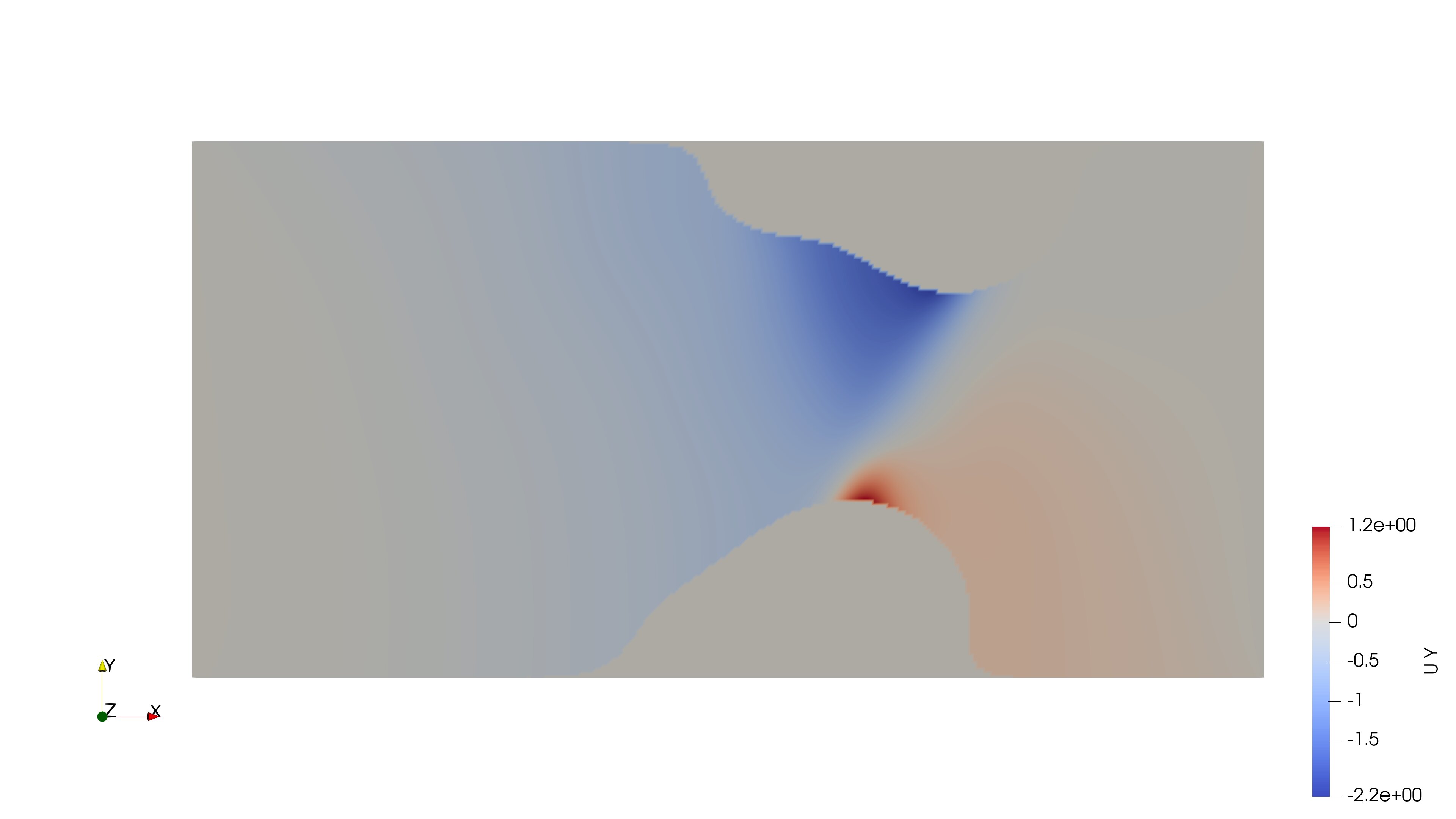} & \includegraphics[width=\widthfigures, valign=c]{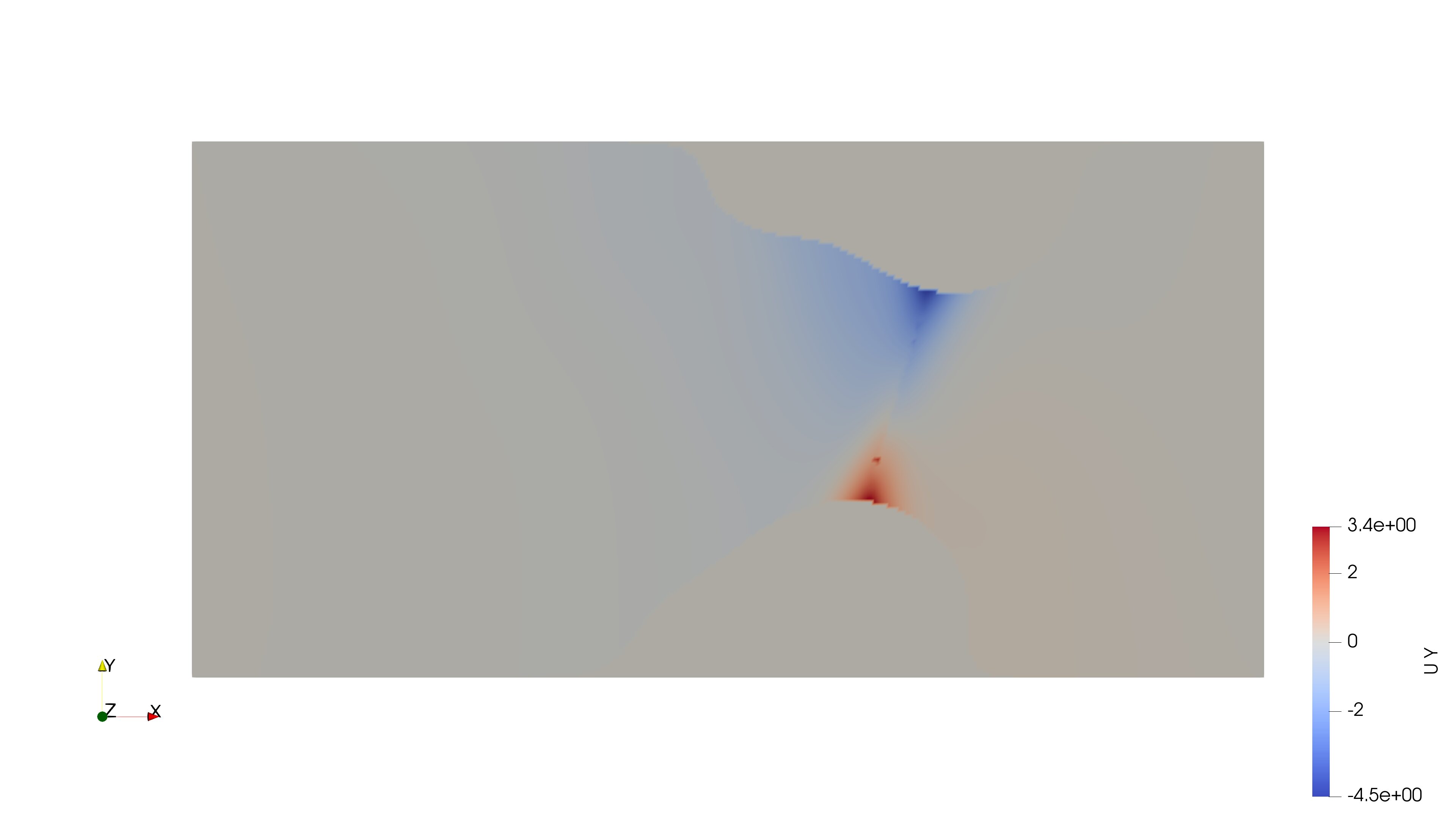} \\
\end{tblr}
  \label{tab:learning_features}
\end{table}

The training was parallelized on 40 GPUs (A100) and lasted 6 hours. Inference and testing can be performed on a single GPU to compute the metrics presented in Table~\ref{tab:PLAID_benchmark}.

\subsubsection{Other datasets}

For the remaining datasets, which correspond to stationary problems, we train FNO models to directly map input quantities to output fields without temporal recursion.

The input features depend on each dataset and typically include scalar parameters (e.g., boundary conditions or material coefficients) together with spatial coordinate fields. When field inputs are available, they are projected onto a regular grid and concatenated as additional channels.

Similarly to the transient case, all data defined on unstructured meshes are projected onto regular grids prior to training. Predictions produced by the FNO are then mapped back onto the original meshes for evaluation using the PLAID benchmark metrics.

\paragraph{Model and training parametrization.}
We summarize all the models parametrization in Table~\ref{tab:param_fno}.

\begin{table}[!htbp]
\small
\centering
  \caption{FNO hyperparameters and training statistics.}
 \begin{tabularx}{\textwidth}{L{2.75cm} C{1.8cm} C{1.1cm} C{1.3cm} C{1.1cm} C{1.3cm} C{1.1cm} C{1.1cm} C{2cm}}
  \toprule
 \texttt{Dataset}                       & Grid shape & Layer count & Latent channels & Decoder layers  & Decoder layers size & Fourier modes   & Training time     & Hardware\\
  \midrule
  \small{\texttt{Tensile2d}}            & $201^2$ & 4           & 64        & 4  &  64  & $16^2$   & 2h46     & 1 $\times$ A30 \\
  \small{\texttt{2D\_MultiScHypEl}}     & $201^2$  & 4          & 64        & 4  &  64  & $16^2$   & 4h12     & 1 $\times$ A30 \\
  \small{\texttt{Rotor37}}              & $64^3$  & 4          & $32$      & 4  &  32  & $32^2$   & 7h19     & 1 $\times$ A30 \\
  \small{\texttt{2D\_profile}}          & 601 $\times$ 801 & 4           & 32        & 2  &  32  & $32^2$   & 2h28     & 1 $\times$ A30 \\
  \small{\texttt{VKI-LS59}}             & 301 $\times$ 121  & 4          & 64        & 4  &  64  & $16^2$   & 2h36     & 1 $\times$ A30 \\
  \small{\texttt{2D\_ElPlDynamics} }    & 301 $\times$ 151  & 4          & 64        & 1  &  64  & $20^2$   & 6h       & 40 $\times$ A100 \\
  \bottomrule
 \end{tabularx}
 \label{tab:param_fno}
\end{table}


\subsection{MARIO}

Modulated Aerodynamic Resolution-Invariant Operator (MARIO) leverages implicit neural representations, allowing continuous modeling of fields without relying on fixed discretizations, which improves robustness to heterogeneous geometries.

MARIO is a deep learning model designed to approximate the solution operator of a partial differential equation (PDE)~\cite{mario_paper}, involving geometric variability. It leverages Conditional Neural Fields (or Implicit Neural Representations) to learn the mapping between spatial coordinates from a mesh, geometric information (e.g., via the signed distance function, SDF), inflow conditions, and the resulting physical field. Unlike mesh-based methods, INRs represent continuous fields through neural network parameterizations, enabling resolution-independent predictions and flexible evaluation. MARIO extends this approach to handle multiple geometries and operating conditions through a conditioning mechanism.

\subsubsection{Method}

\paragraph{Modulated INR architecture.}
MARIO implements a conditional neural field approach where a single neural network architecture can represent multiple distinct signals through a conditioning mechanism. The conditioning variable $z = [\mu_{\text{geom}}, \mu]$ encodes both geometric parameterization $\mu_{\text{geom}}$ and operating conditions $\mu$ (e.g., angle of attack, Mach number, Reynolds number).

The main network is a multilayer perceptron (MLP) where the layer outputs are modulated by sample-specific vectors:

\begin{align}
f_{\theta,\phi}(x) &= W_L(\eta_{L-1} \circ \eta_{L-2} \circ \cdots \circ \eta_1 \circ \gamma(x)) + b_L \\
\eta_l(\cdot) &= \text{ReLU}(W_l(\cdot) + b_l + \phi_l(z))
\end{align}

where $\phi_l(z) = [h_{\psi}(z)]_l \in \mathbb{R}^{d_l}$ are layer-specific modulation vectors obtained from the hypernetwork $h_{\psi}$ that processes the conditioning variable $z$. The main network parameters $\theta$ are shared for all samples and consist of the weights and biases matrices $W_l, b_l$. In MARIO, an explicit shape encoding $\mu_{\text{geom}}$ is used as input of the architecture to properly model geometric variability. In many real-world applications, a geometric parameterization is not available or insufficient to capture complex shapes. Therefore, a learning mechanism to obtain compact geometric representations from the SDF fields is adopted. This encoding process leverages a separate Neural Field encoder, that maps input coordinates to output SDF values, while fitting latent shape representations.

\paragraph{Geometry encoding mechanism.}
For each geometry's signed distance function (SDF), a meta-learning optimization procedure based on CAVIA \cite{zintgraf2019fast} adapts a shared neural network $f_{\theta_{in},\phi_{in}}$ to represent different shapes. Given the shared network parameters $\theta_{in}$ and hypernetwork parameters $\psi$, the latent representation $\mu_{geom}=z_{in}^{(K)}$ for geometry $i$ is obtained by solving:

\begin{align}
    z_{in}^{(0)} &= 0 \\
    z_{in}^{(k+1)} &= z_{in}^{(k)} - \alpha \nabla_{z_{in}^{(k)}} \mathcal{L}_{in}(f_{\theta_{in},\phi_{in}}(x), sdf_i), \quad \text{for } 0 \leq k \leq K-1
\end{align}

where $\phi_{in}=h_{\psi}(z_{in}^{(k)})$, $\alpha$ is the inner loop learning rate, and $K$ is the number of optimization steps (typically set to 3). The loss $\mathcal{L}_{in}$ measures the reconstruction error between the true SDF field and its prediction over a sampling grid defined on the input domain.

\begin{figure}
    \centering
    \includegraphics[width=0.8\linewidth]{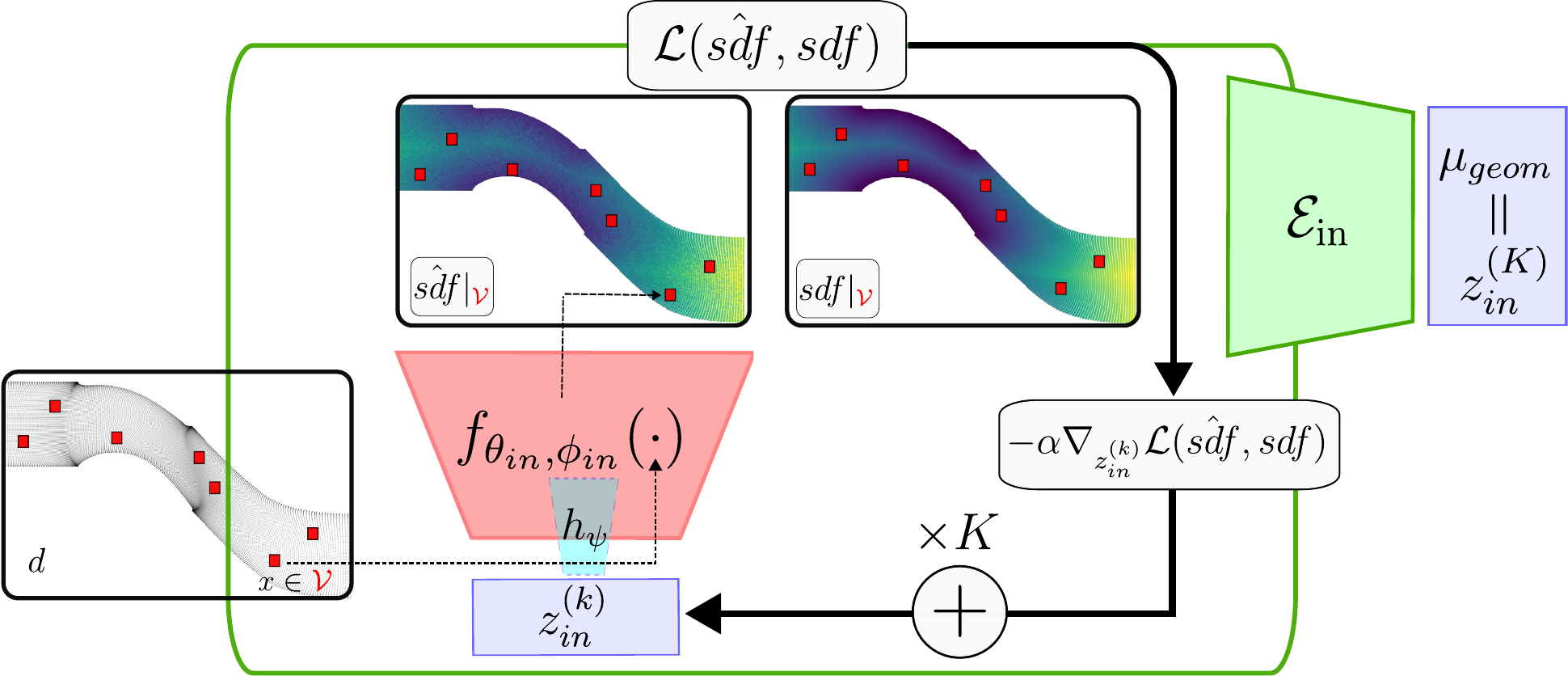}
    \caption{MARIO geometry encoding process.}
    \label{fig:MARIO_encoding}
\end{figure}

This optimization process, illustrated in Figure~\ref{fig:MARIO_encoding}, yields a compact latent code $\mu_{\text{geom}} = z_{in}^{(K)}$ that captures the essential geometric features.

\paragraph{Fourier feature encoding.}
To address the spectral bias inherent in neural networks, MARIO employs Fourier feature encoding for the input coordinates:

\begin{equation}
\gamma(x) = [\cos(2\pi \mathbf{B}x), \sin(2\pi \mathbf{B}x)]
\end{equation}

where $\mathbf{B} \in \mathbb{R}^{m \times d}$ contains frequency vectors sampled from a Gaussian distribution $\mathcal{N}(0,\sigma)$. This encoding enables the network to better capture high-frequency details in the output fields.

\paragraph{Scalar output prediction.}
In addition to predicting coordinate-dependent fields, MARIO can also predict global scalar quantities for each sample. Since these scalar outputs are global properties of the solution (e.g., power coefficients, efficiency metrics), they depend only on the sample-specific information encoded in the modulation vectors. The scalar prediction is therefore implemented as:

\begin{equation}
s = W_s \cdot \phi_{agg} + b_s
\end{equation}

where $\phi_{agg}$ represents an aggregation of the modulation vectors produced by the hypernetwork. This single-layer transformation efficiently leverages the already learned sample representation without requiring additional feature extraction.

\begin{figure}
    \centering
    \includegraphics[width=0.8\linewidth]{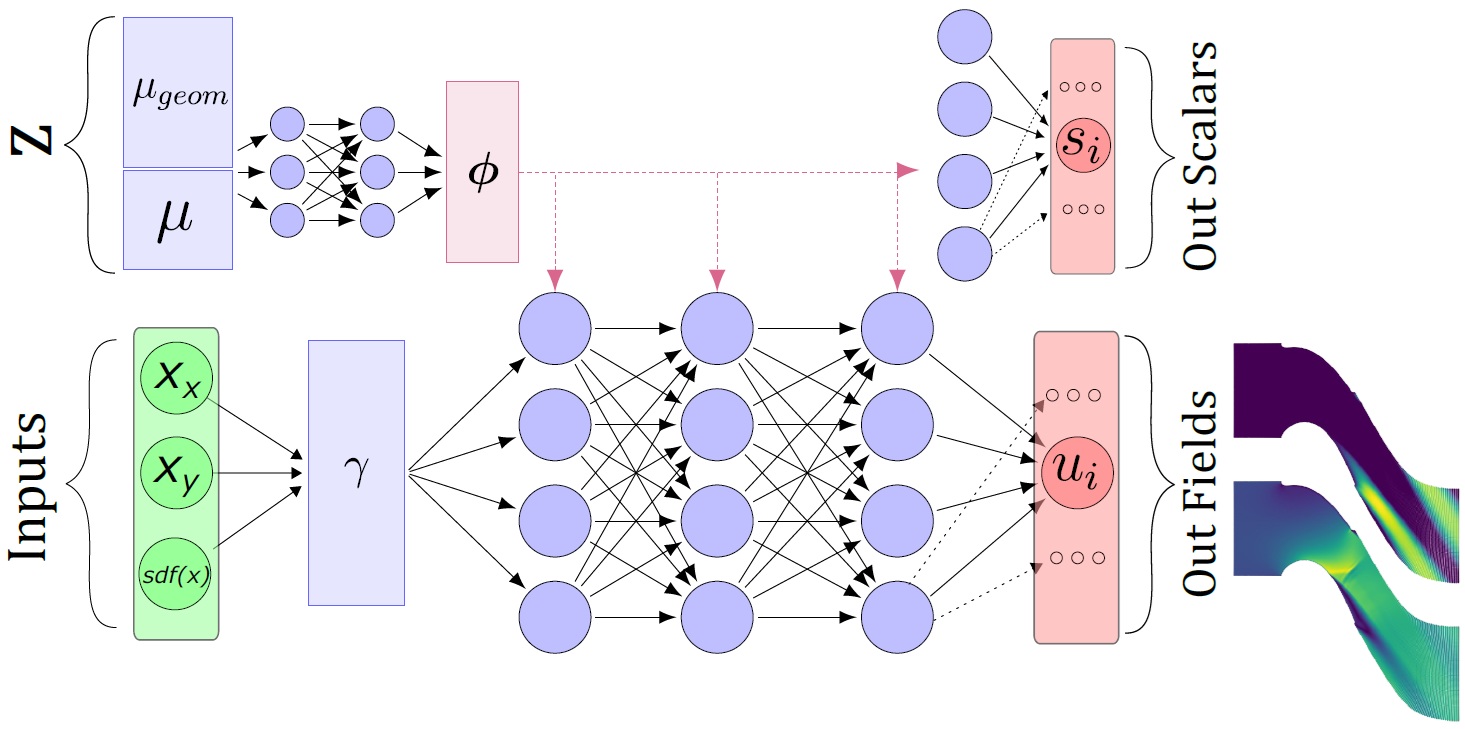}
    \caption{MARIO architecture.}
    \label{fig:MARIO_architecture}
\end{figure}

The architecture of MARIO is illustrated in Figure~\ref{fig:MARIO_architecture}.

\paragraph{Training procedure.}
MARIO is trained using a weighted loss function that balances field prediction accuracy and scalar output accuracy:

\begin{equation}
\mathcal{L} = \alpha \cdot \mathcal{L}_{\text{field}} + (1-\alpha) \cdot \mathcal{L}_{\text{scalar}}
\end{equation}

where $\alpha \in [0,1]$ is a weighting parameter. The field loss $\mathcal{L}_{\text{field}}$ is computed as the mean squared error between predicted and target fields across spatial locations, while the scalar loss $\mathcal{L}_{\text{scalar}}$ is the mean squared error of the global quantities.

\paragraph{Key advantages.}
MARIO presents three major benefits: (i) it is resolution-invariant and can be evaluated at arbitrary spatial locations; (ii) it overcomes spectral bias through multiscale Fourier encodings; and (iii) it adapts to geometry-specific variations via bias modulation using the auxiliary network $h_\psi$.

\subsubsection{Experiment}

The MARIO model is applied to the six datasets, with the parametrization and training procedures provided in Table~\ref{tab:MARIO_param1}.


\begin{table}[h!]
\small
\centering
  \caption{MARIO hyperparameters and training statistics.}
 \begin{tabularx}{\textwidth}{L{2.6cm} C{1cm} C{1cm} C{1cm} C{1cm} C{1cm} C{0.9cm} C{0.9cm} C{0.9cm} C{0.9cm} C{1.4cm}}
  \toprule
 \texttt{Dataset}  &{Geom. Hyper. depth} & {Geom. Hyper. width} & {Geom. latent dim} & {Hypernet. depth} & {Hypernet. width} & {INR depth} & {INR width} & {freq. nb.}  & Training time     & Hardware\\
  \midrule
  \small{\texttt{Tensile2d}}            & 1 & 128 & 16 &  3 & 256 & 6  &  256  & 64 & 3h40  & 1 $\times$ A30\\
  \small{\texttt{2D\_MultiScHypEl}}     & 1 & 128 & 100 & 3 & 256 & 6  &  256  & 64 & 6h52  & 1 $\times$ A30\\
  \small{\texttt{Rotor37}}              & 1 & 128 & 32 & 3 & 256 & 6  &  256  & 64  & 4h45  & 1 $\times$ A100\\
  \small{\texttt{2D\_profile}}          & 1 & 128 & 8  & 3 & 256 & 6  &  256  & 64  & 2h43  & 1 $\times$ A100\\
  \small{\texttt{VKI-LS59}}             & 1 & 128 & 16 & 3 & 256 & 6  &  256  & 64  & 8h45 & 1 $\times$ A100 \\
  \small{\texttt{2D\_ElPlDynamics} }    & 1 & 128 & 16 & 3 & 256 & 6  &  256  & 64  & 26min  & 1 $\times$ A30\\
  \bottomrule
 \end{tabularx}
 \label{tab:MARIO_param1}
\end{table}

\section{Additional details on PLAID}
\label{app:plaid}

We illustrate further the capabilities of PLAID by providing some additional commands to retrieve information from our datasets directly from Hugging Face.

\subsection{\texttt{Tensile2d}}

\texttt{Tensile2d} is a simple dataset, for which standard and simple PLAID commands are sufficient to retrieve the data:
\begin{python}
from datasets import load_dataset
from plaid.containers.sample import Sample
import pickle

# Load the dataset
hf_dataset = load_dataset("PLAID-datasets/Tensile2d", split="all_samples")

# Get split ids
ids_train = hf_dataset.description["split"]["train_500"]

# Get inputs/outputs names
in_scalars_names = hf_dataset.description["in_scalars_names"]
out_fields_names = hf_dataset.description["out_fields_names"]

# Get samples
sample = Sample.model_validate(pickle.loads(hf_dataset[ids_train[0]]["sample"]))

# Examples of data retrievals
nodes = sample.get_nodes()
elements = sample.get_elements()
nodal_tags = sample.get_nodal_tags()

for sn in ["P", "p1", "p2", "p3", "p4", "p5"]:
    scalar = sample.get_scalar(sn)

# outputs
for fn in ["U1", "U2", "q", "sig11", "sig22", "sig12"]:
    field = sample.get_field(fn)

for sn in ["max_von_mises", "max_q", "max_U2_top", "max_sig22_top"]:
    scalar = sample.get_scalar(sn)

\end{python}

The geometrical support in PLAID samples can be easily converted to Muscat meshes:
\begin{python}
from Muscat.Bridges import CGNSBridge
CGNS_tree = sample.get_tree()
mesh = CGNSBridge.CGNSToMesh(CGNS_tree)
\end{python}

\subsection{\texttt{VKI-LS59}}
\label{app:plaid-vki}

\begin{figure}
    \centering
    \includegraphics[width=0.8\linewidth]{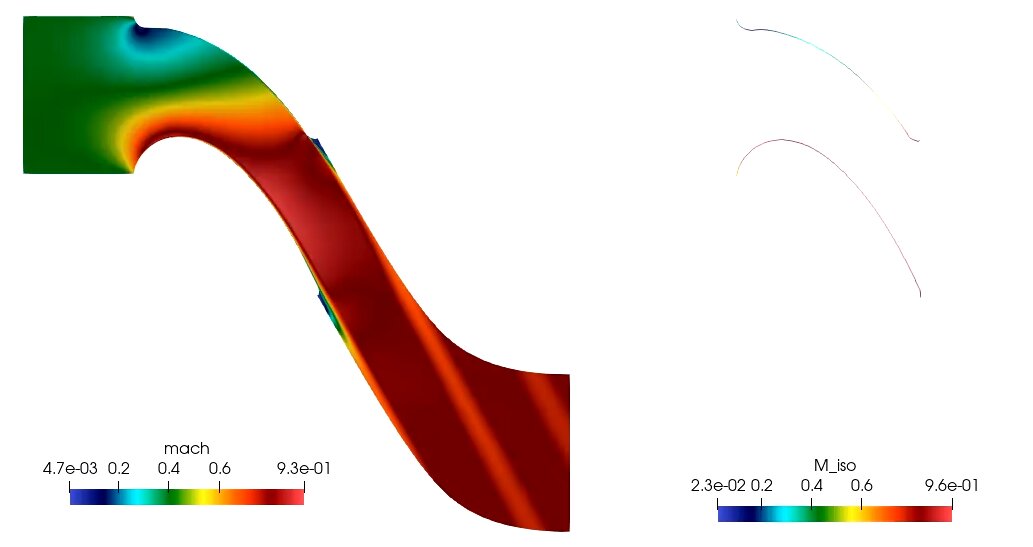}
    \caption{Illustration of the first sample in the train split of \texttt{VKI-LS59}: (left) fluid domain, (right) blade surface domain.}
    \label{fig:vki_geom}
\end{figure}

\texttt{VKI-LS59} also contains stationary configurations, meaning only one time step per sample, but features a complex geometrical setting, with a 2D fluid domain and a 1D blade surface domain, see Figure~\ref{fig:vki_geom}.

The fluid domain contains 2D elements in a 2D ambient space, hence is contained in the CGNS base called "Base\_2\_2".
For the blade surface domain, we have 1D elements in a 2D ambient space: the CGNS base is then "Base\_1\_2". The corresponding data are retrieved as follows:
\begin{python}
from datasets import load_dataset
from plaid.containers.sample import Sample
import pickle

# Load the first sample of the train split
hf_dataset = load_dataset("PLAID-datasets/VKI-LS59", split="all_samples")
ids_train = hf_dataset.description["split"]["train"]
sample = Sample.model_validate(pickle.loads(hf_dataset[ids_train[0]]["sample"]))

# Examples of data retrievals
for fn in ["sdf", "ro", "rou", "rov", "roe", "nut", "mach"]:
    field = sample.get_field(fn, base_name="Base_2_2")
M_iso = sample.get_field("M_iso", base_name="Base_1_2")
for sn in sample.get_scalar_names():
    scalar = sample.get_scalar(sn)

nodes_fluid = sample.get_nodes(base_name="Base_2_2")
nodes_blade_surface = sample.get_nodes(base_name="Base_1_2")
elements_fluid = sample.get_elements(base_name="Base_2_2")
elements_blade_surface = sample.get_elements(base_name="Base_1_2")
nodal_tag_fluid = sample.get_nodal_tags(base_name="Base_2_2")
\end{python}

The meshes for the fluid domain and blade surface domain can also be converted to Muscat meshes:
\begin{python}
from Muscat.Bridges import CGNSBridge
CGNS_tree = sample.get_tree()
mesh_fluid = CGNSBridge.CGNSToMesh(CGNS_tree, baseNames=["Base_2_2"])
mesh_blade = CGNSBridge.CGNSToMesh(CGNS_tree, baseNames=["Base_1_2"])
\end{python}

\subsection{\texttt{2D\_ElPlDynamics}}
\label{app:plaid-plastodyn}

\texttt{2D\_ElPlDynamics} contains additional complexity: time-dependent data and a field located at the center of the elements. When retrieving data, the default location of the fields is at the vertices. For other type of fields, the location must be specified.
The corresponding commands are provided below:

\begin{python}
from datasets import load_dataset
from plaid.containers.sample import Sample
import pickle

# Load the first sample of the train split
hf_dataset = load_dataset("PLAID-datasets/2D_ElastoPlastoDynamics", split="all_samples")
ids_train = hf_dataset.description["split"]["train"]
sample = Sample.model_validate(pickle.loads(hf_dataset[ids_train[0]]["sample"]))

# Examples of data retrievals
time_steps = sample.get_all_time_values()

for time in time_steps:
    for fn in ["U_x","U_y"]:
        field = sample.get_field(fn, time = time)
    field = sample.get_field("EROSION_STATUS", location="CellCenter", time = time)

CGNS_tree = sample.get_tree(time = 0.1)
\end{python}

\section{Benchmarking online applications}
\label{app:bench}

Anyone wishing to participate in our benchmarks, hosted at \href{https://huggingface.co/PLAIDcompetitions}{huggingface.co/PLAIDcompetitions}, should create a Hugging Face account. However, no account is required to browse the website or view the leaderboards. To participate, users simply need to train their model independently and submit predictions on the testing set. We do not require participants to upload their models. Two separate leaderboards are maintained, each based on a hidden subset of the test set, in order to discourage attempts to overfit on the testing set.

We illustrate the benchmarking application using the \texttt{VKI-LS59} dataset as an example.

\begin{figure}[h!]
    \centering
    \includegraphics[width=\linewidth]{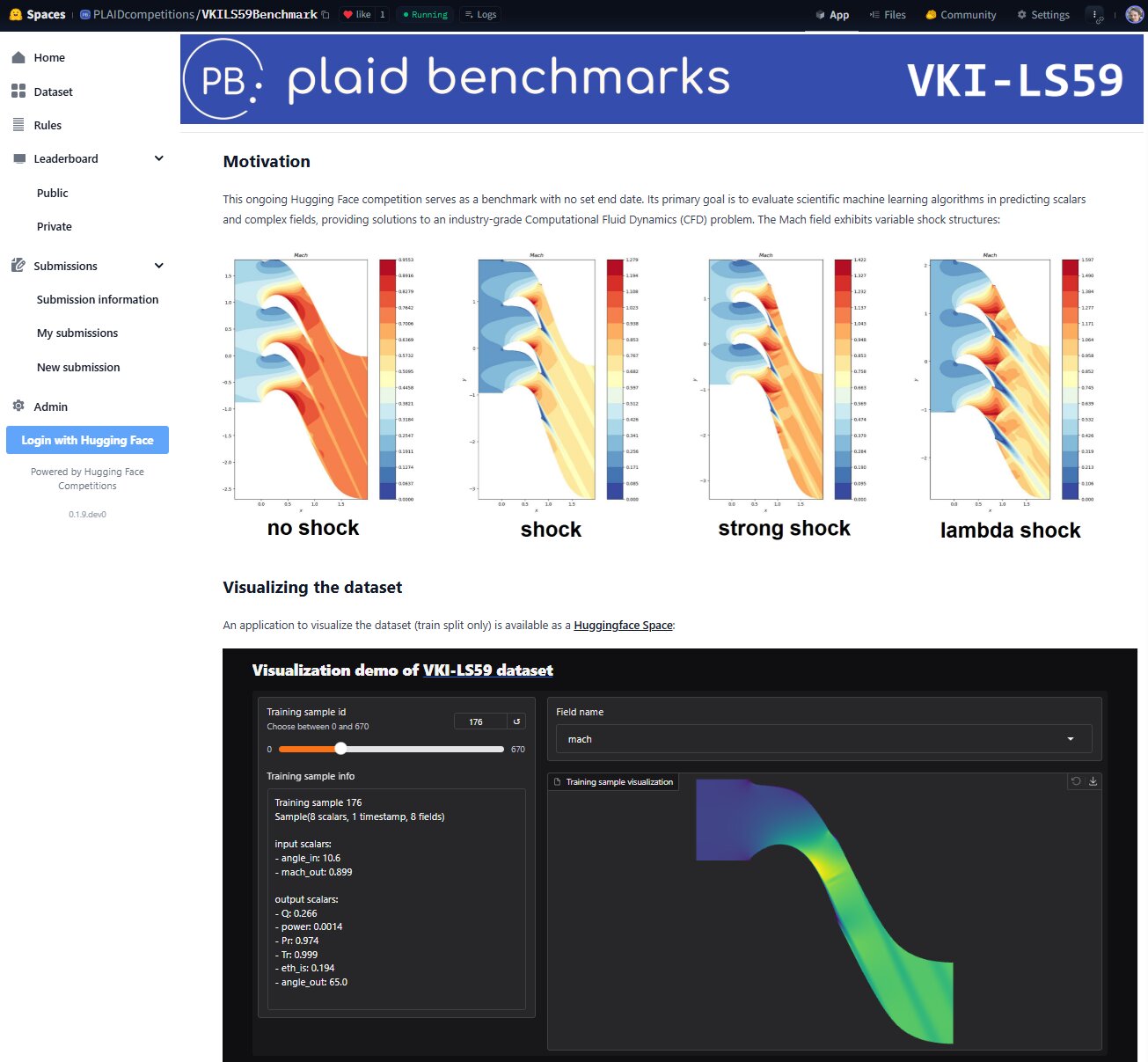}
    \caption{"Home" page of the benchmarking application on the \texttt{VKI-LS59} dataset.}
    \label{fig:bench_app_homepage}
\end{figure}

Figure~\ref{fig:bench_app_homepage} shows the benchmark homepage. A navigation menu is available on the left-hand side, allowing users to browse the site and log in. This page also provides examples of the dataset output fields and includes a visualization tool, where users can select a training sample ID and an output field to display.

\begin{figure}[h!]
    \centering
    \includegraphics[width=0.8\linewidth]{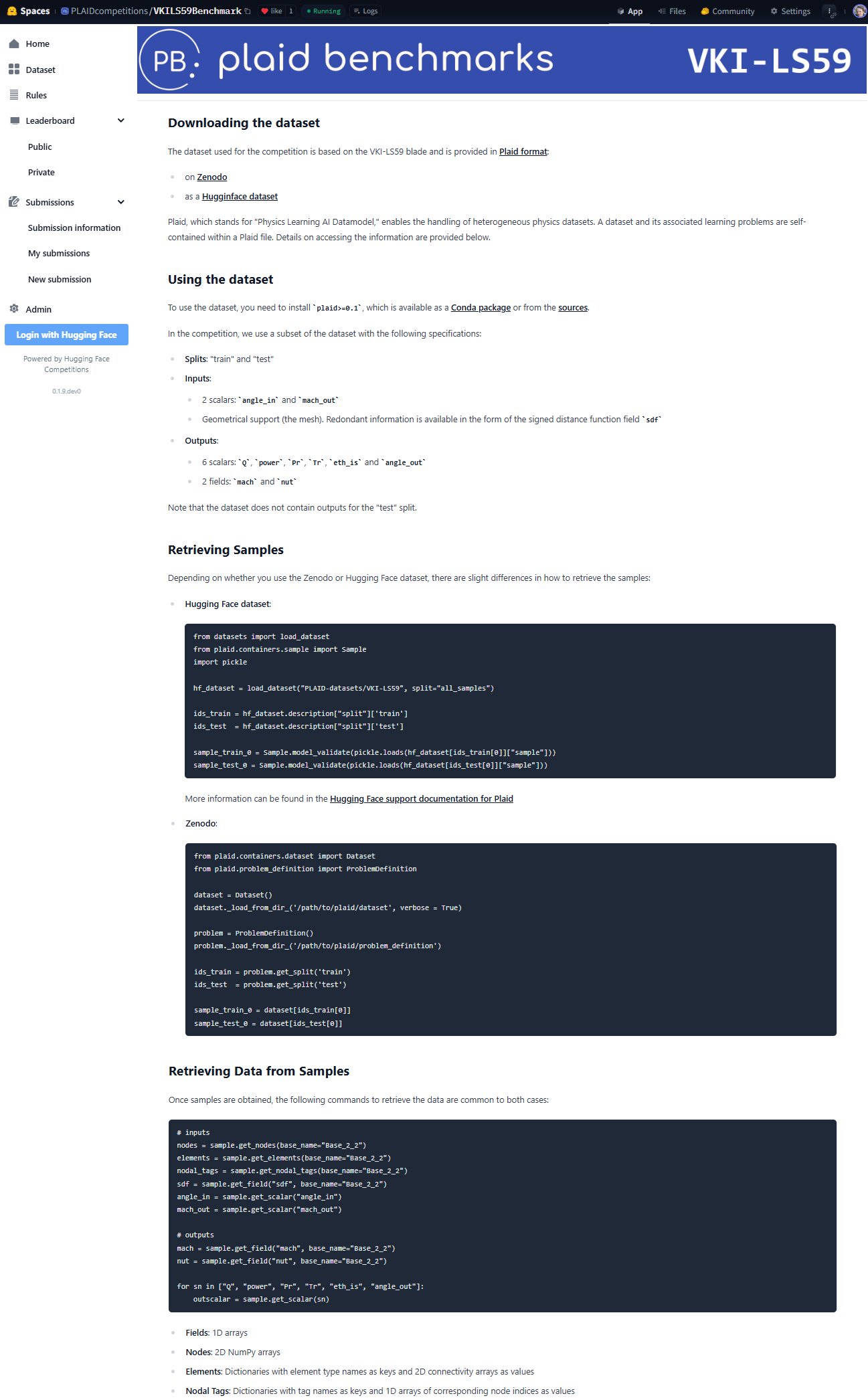}
    \caption{"Dataset" page of the benchmarking application on the \texttt{VKI-LS59} dataset.}
    \label{fig:bench_app_dataset}
\end{figure}

Figure~\ref{fig:bench_app_dataset} provides detailed instructions on how to retrieve the dataset, including a description of the inputs and outputs used in the benchmark. Example commands are also provided to retrieve the samples and the required associated data.

\begin{figure}[h!]
    \centering
    \includegraphics[width=\linewidth]{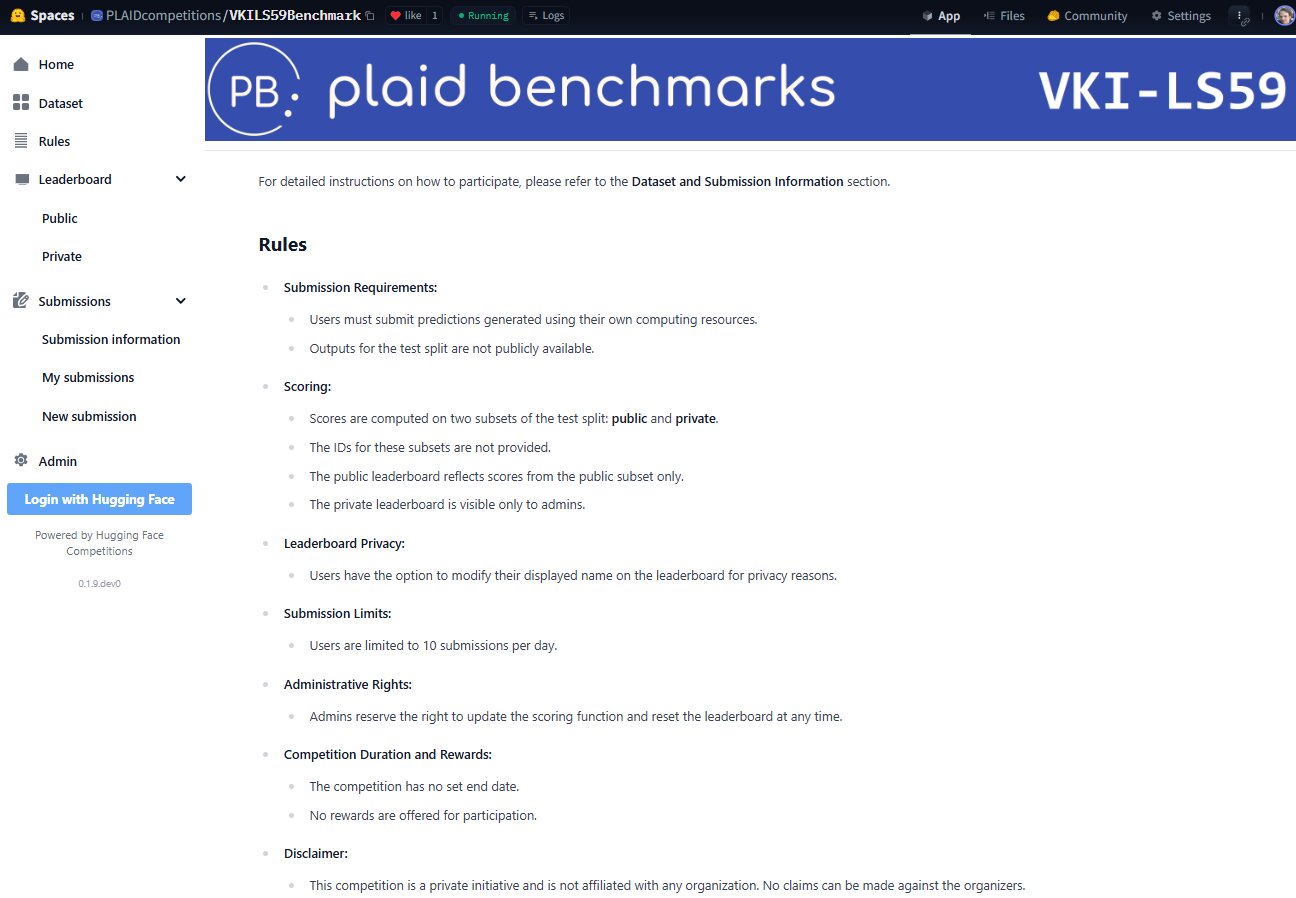}
    \caption{"Rules" page of the benchmarking application on the \texttt{VKI-LS59} dataset.}
    \label{fig:bench_app_rules}
\end{figure}

The set of rules applying to the benchmark is presented in Figure~\ref{fig:bench_app_rules}.

\begin{figure}[h!]
    \centering
    \includegraphics[width=\linewidth]{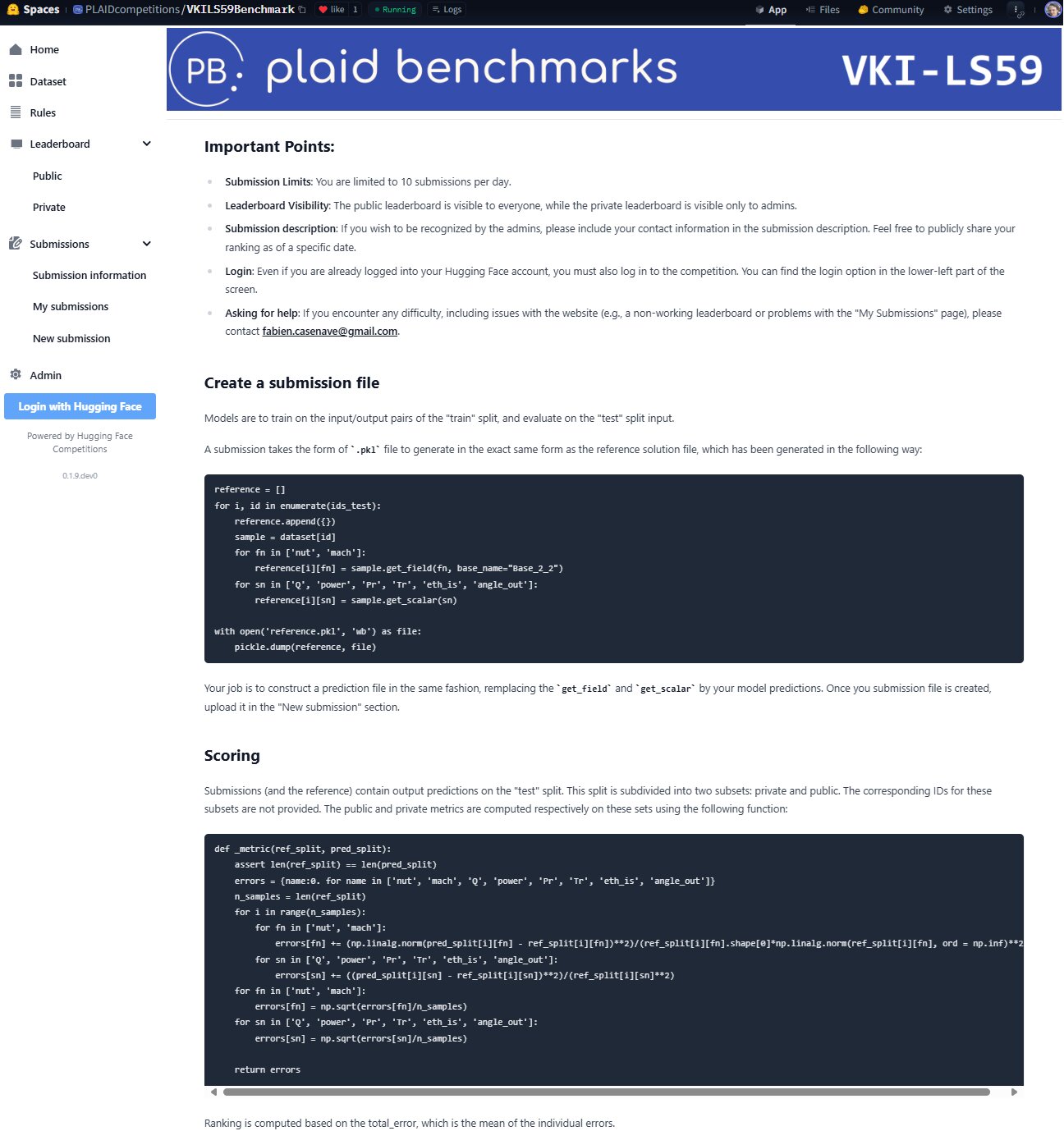}
    \caption{"Submission information" page of the benchmarking online application on the \texttt{VKI-LS59} dataset.}
    \label{fig:bench_app_submission_info}
\end{figure}

Figure~\ref{fig:bench_app_submission_info} provides detailed instructions on how to generate and submit the prediction file. The scoring function used for evaluation is also described.

\begin{figure}[h!]
    \centering
    \includegraphics[width=\linewidth]{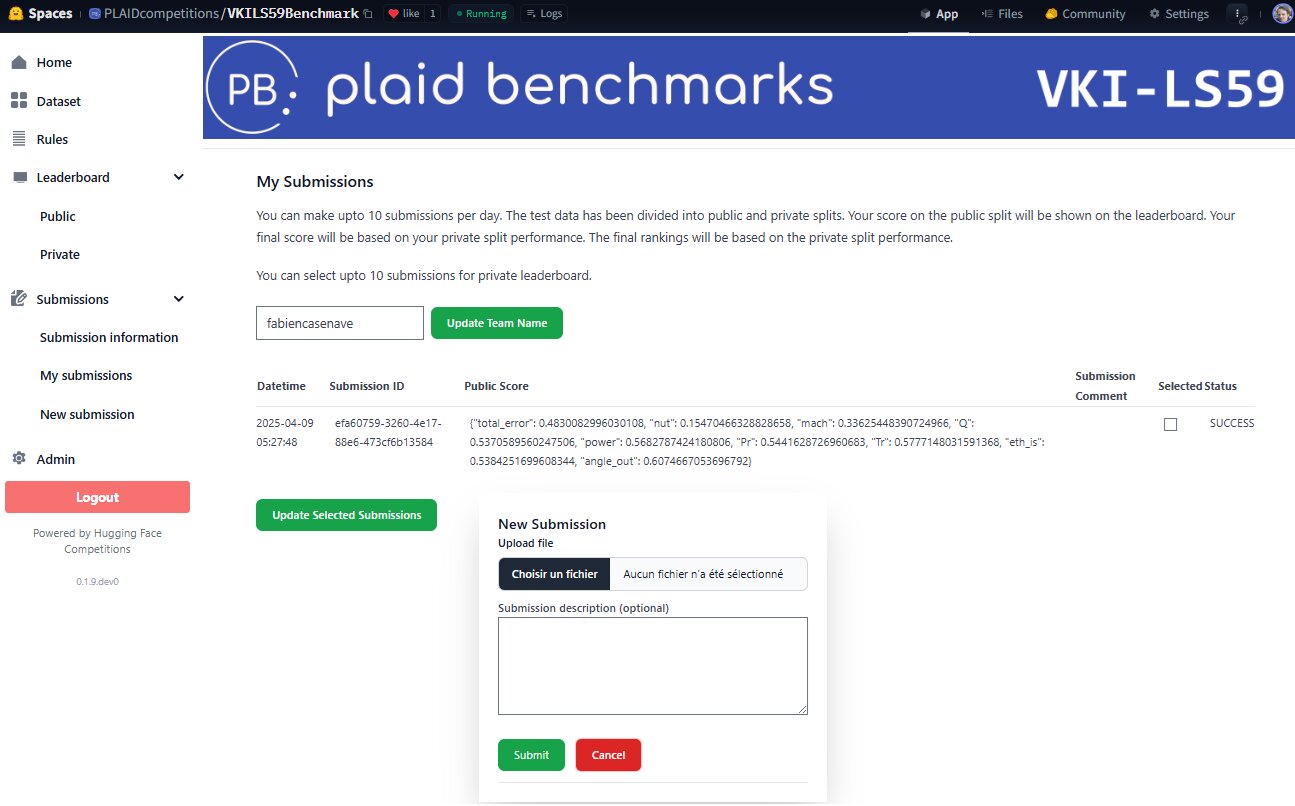}
    \caption{"My submissions" page of the benchmarking application on the \texttt{VKI-LS59} dataset.}
    \label{fig:bench_app_submission_example}
\end{figure}

Figure~\ref{fig:bench_app_submission_example} illustrates the user’s submissions page and the submission interface.

\begin{figure}[h!]
    \centering
    \includegraphics[width=\linewidth]{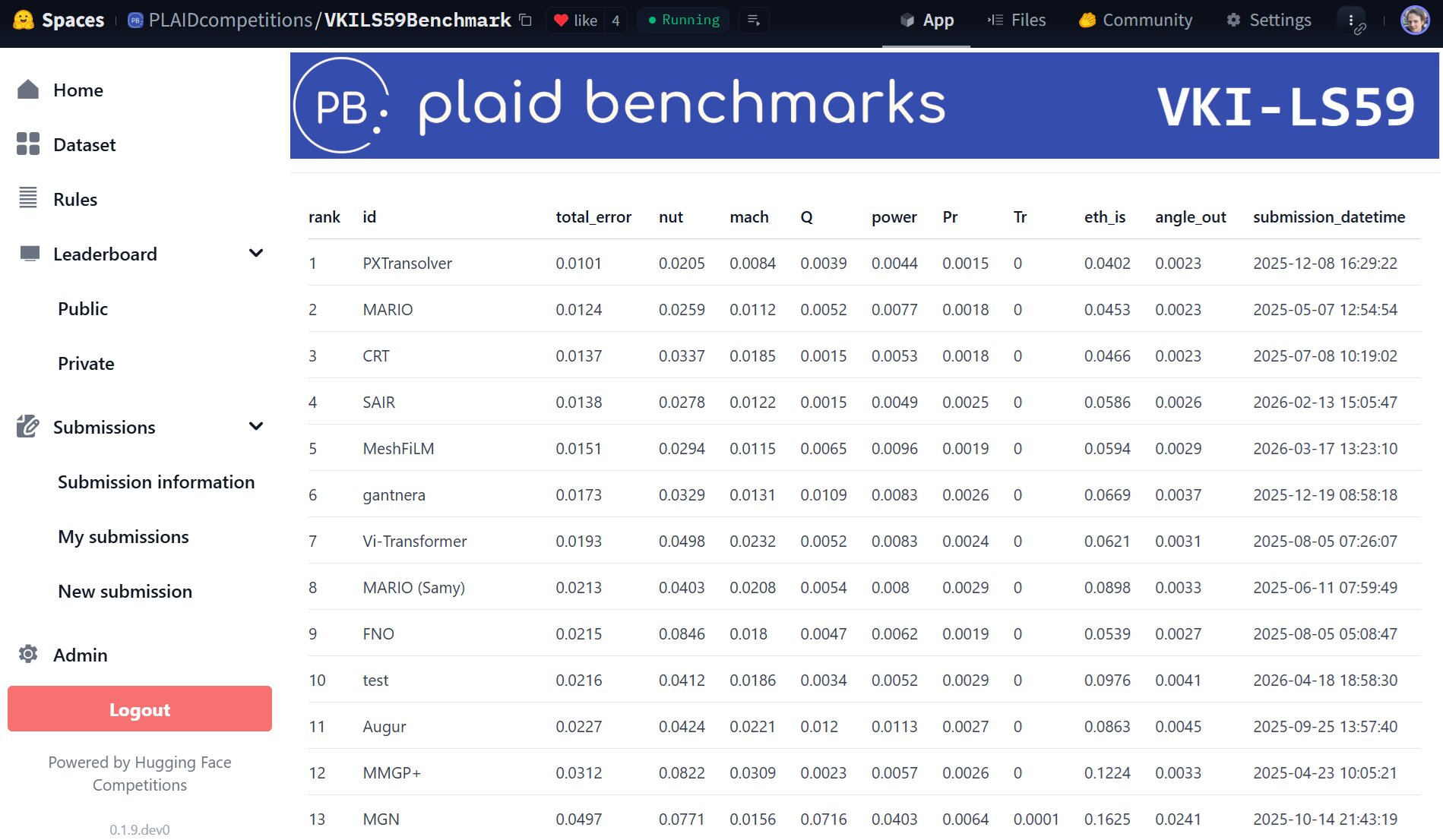}
    \caption{“Public leaderboard” page of the benchmarking application for the \texttt{VKI-LS59} dataset, as of April 19, 2026. Additional leaderboard entries have been submitted by external contributors.}
    \label{fig:bench_app_leaderboard}
\end{figure}

Figure~\ref{fig:bench_app_leaderboard} shows the public leaderboard as it appeared at the time of submission of this work.

\end{document}